\documentclass[9pt, conference]{IEEEtran}
\usepackage{cite}
\usepackage{amsmath,amssymb,amsfonts}
\usepackage{graphicx}
\usepackage{textcomp}
\usepackage{xcolor}
\usepackage{soul}
\usepackage{subfigure}
\usepackage{booktabs}
\usepackage{tabularx, tabu}
\usepackage{xspace}
\usepackage{mathtools}
\usepackage{caption}
\usepackage[colorlinks]{hyperref}
\usepackage{multirow}
\usepackage{float}

\newcommand*\rfrac[2]{{}^{#1}\!/_{#2}}

\usepackage{tikz}
\newcommand*\circled[1]{\tikz[baseline=(char.base)]{
            \node[shape=circle,fill,inner sep=0.8pt] (char) {\textcolor{white}{#1}};}}

\newcommand\wcircled[1]{\tikz[baseline=(char.base)]{ 
        \node (X) [shape=circle, draw, inner sep=0.8pt] (char) 
        {\textcolor{black}{#1}};}}

\usepackage[a4paper, total={184mm,239mm}]{geometry}
\def\BibTeX{{\rm B\kern-.05em{\sc i\kern-.025em b}\kern-.08em
    T\kern-.1667em\lower.7ex\hbox{E}\kern-.125emX}}
    
\usepackage[ruled,linesnumbered]{algorithm2e}
\newlength\mylenin
\newcommand\myinput[1]{%
\settowidth\mylenin{\KwIn{}}%
\setlength\hangindent{\mylenin}%
\hspace*{\mylenin}#1\\}

\let\oldnl\nl
\newcommand{\nonl}{\renewcommand{\nl}{\let\nl\oldnl}}

\newlength\mylenout

\usepackage{algpseudocode}

\usepackage{tikz}

\makeatletter
\def\footnoterule{\relax%
  \kern-5pt
  \hbox to \columnwidth{\hfill\vrule width 1\columnwidth height 0.4pt\hfill}
  \kern4.6pt}
\makeatother

\definecolor{stelios_colour}{RGB}{200, 238, 200}
\definecolor{light_red}{RGB}{255, 204, 204}

\definecolor{crimson}{rgb}{0.86, 0.08, 0.24}

\newif\ifcomment

\commentfalse

\ifcomment
\newcommand{\stelios}[1]{\sethlcolor{stelios_colour}\hl{[\textbf{Stelios:} #1]}}
\newcommand{\steve}[1]{\sethlcolor{cyan}\hl{[Steve: #1]}}
\newcommand{\nic}[1]{\sethlcolor{yellow}\hl{[Nic: #1]}}
\newcommand{\alex}[1]{\sethlcolor{orange}\hl{[Alex: #1]}}
\newcommand{\cut}[1]{\sethlcolor{light_red}\hl{[#1]}}
\newcommand{\blue}[1]{\textcolor{blue}{#1}}
\else
\newcommand{\stelios}[1]{}
\newcommand{\steve}[1]{}
\newcommand{\alex}[1]{}
\newcommand{\nic}[1]{}
\newcommand{\cut}[1]{}
\newcommand{\blue}[1]{\textcolor{black}{#1}}
\fi

\newcommand\blfootnote[1]{%
  \begingroup
  \renewcommand\thefootnote{}\footnote{#1}%
  \addtocounter{footnote}{-1}%
  \endgroup
}

\begin{document}




\title{Fluid Batching: Exit-Aware Preemptive Serving of \\ Early-Exit Neural Networks on Edge NPUs}

\author{
\IEEEauthorblockN{Alexandros Kouris$^{*}$,\hspace{1mm} Stylianos I. Venieris$^{*}$}
\IEEEauthorblockA{\small\textit{Samsung AI Center, Cambridge, UK}\\
{\small \{a.kouris, s.venieris\}@samsung.com}}
\and
\IEEEauthorblockN{Stefanos Laskaridis$^\dagger$}
\IEEEauthorblockA{\small\textit{Brave Software}\\
{\small  mail@stefanos.cc}}
\and
\IEEEauthorblockN{Nicholas D. Lane$^{\dagger}$}
\IEEEauthorblockA{\small\textit{University of Cambridge} and \textit{Flower Labs}\\
{\small ndl32@cam.ac.uk }}
}


\maketitle

\begin{abstract}
    With deep neural networks (DNNs) emerging as the backbone in a multitude of computer vision tasks, their adoption in real-world applications broadens continuously. Given the abundance and omnipresence of smart devices in the consumer landscape, ``smart ecosystems'' are being formed where sensing happens concurrently rather than standalone. This is shifting the on-device inference paradigm towards deploying centralised neural processing units (NPUs) at the edge, where multiple devices (\textit{e.g.}~in smart homes or autonomous vehicles) can stream their data for processing with dynamic rates. While this provides enhanced potential for input batching, naive solutions can lead to subpar performance and quality of experience, especially under spiking loads. At the same time, the deployment of dynamic DNNs, comprising stochastic computation graphs (\textit{e.g.} early-exit (EE) models), introduces a new dimension of dynamic behaviour in such systems. In this work, we propose a novel early-exit-aware scheduling algorithm that allows sample preemption at run time, to account for the dynamicity introduced both by the arrival and early-exiting processes. At the same time, we introduce two novel dimensions to the design space of the NPU hardware architecture, namely Fluid Batching and Stackable Processing Elements, that enable run-time adaptability to different batch sizes and significantly improve the NPU utilisation even at small batches. Our evaluation shows that the proposed system achieves an average 1.97$\times$ and 6.7$\times$ improvement over state-of-the-art DNN streaming systems in terms of average latency and tail latency \blue{service-level objective (SLO)} satisfaction, respectively.

\end{abstract}

\blfootnote{${}^*$Equal Contribution.}
\blfootnote{${}^{\dagger}$Work done while with Samsung AI.}

\section{Introduction}

The continued advancement of deep neural networks (DNNs) has led to their mainstream adoption in consumer applications, as a backbone for several computer vision tasks. This has led to the formation of smart ecosystems within the consumer environment, where numerous neighbouring devices are simultaneously collecting an abundance of data. The high concentration of sensing platforms in such local ecosystems has shifted the inference paradigm one step away from the device, at the edge, where data from several sources can be streamed to a more powerful shared compute platform for inference\blue{~\cite{edge_ai_hub2022arxiv}}. Smart homes and mobile robots constitute prime examples of this scenario, as in both cases several visual sensors are continuously capturing images that often need to be processed by the same model (\textit{e.g.}~person identification) under strict latency constraints. 

This increased rate of inference requests on the same model from multiple sources, unlocks the potential of batch processing, which constitutes a promising approach for meeting the throughput demand with a given neural processing unit (NPU) at the edge. By grouping samples and dispatching them together to the NPU, parallelism and data-reuse are increased, leading to improved hardware utilisation, higher processing rate and potentially shorter waiting time for incoming samples. However, batching comes at a cost; the time overhead of assembling samples and the longer computation time induces increased latency which can impact the quality of experience (QoE). As such, the status-quo wisdom dictates that batch processing should be avoided in latency-critical applications. In this context, there is an emerging need for novel methods that leverage the throughput benefits of batching to serve a multitude of inference requests, while also meeting tight latency constraints imposed by the target application so that the QoE is not penalised.

\begin{figure}[t]
    \includegraphics[width=0.49\columnwidth,trim={0cm -1cm 0cm 0cm},clip]{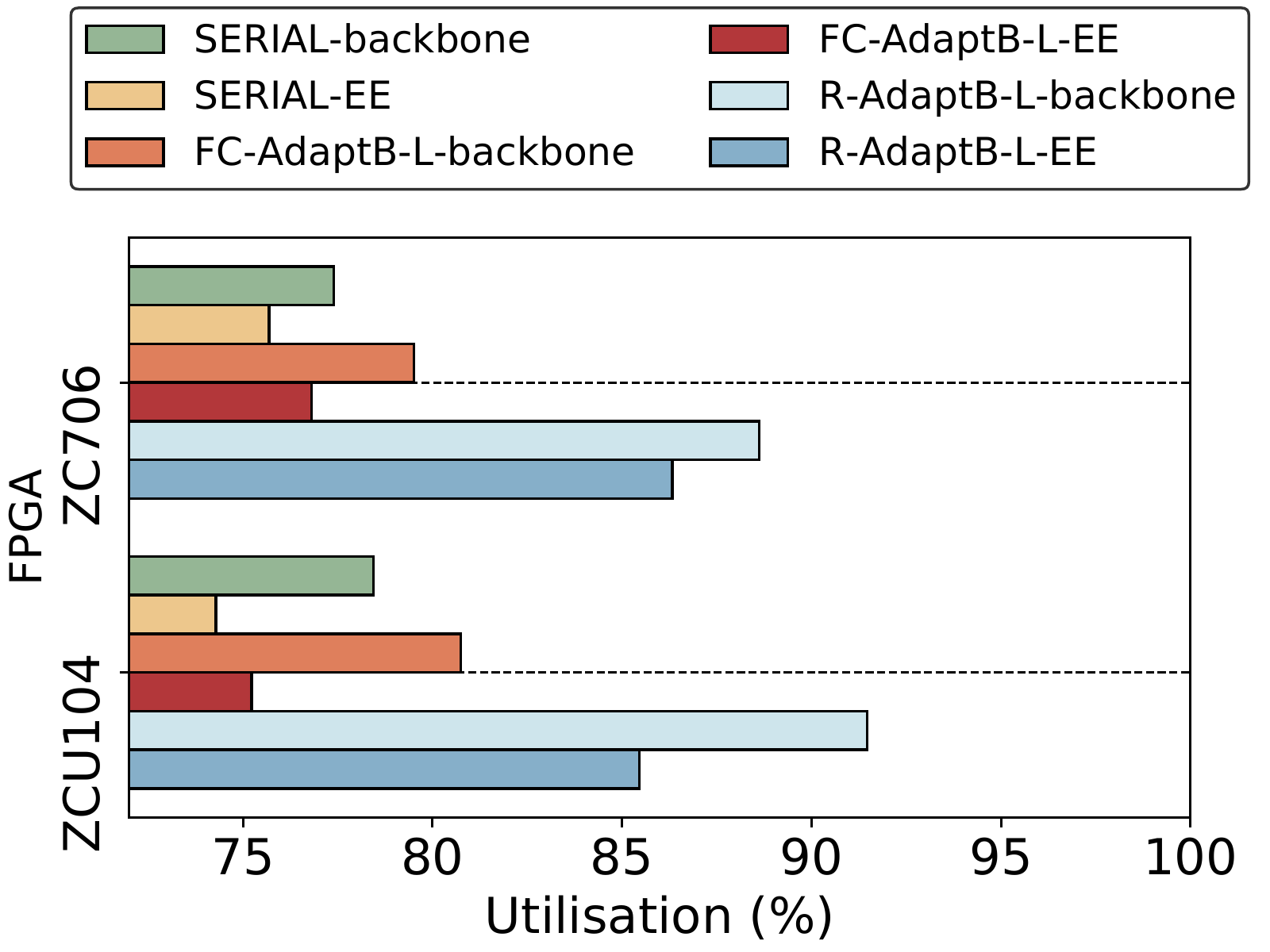} 
    \includegraphics[width=0.49\columnwidth,trim={0cm 0cm 0cm 0cm},clip]{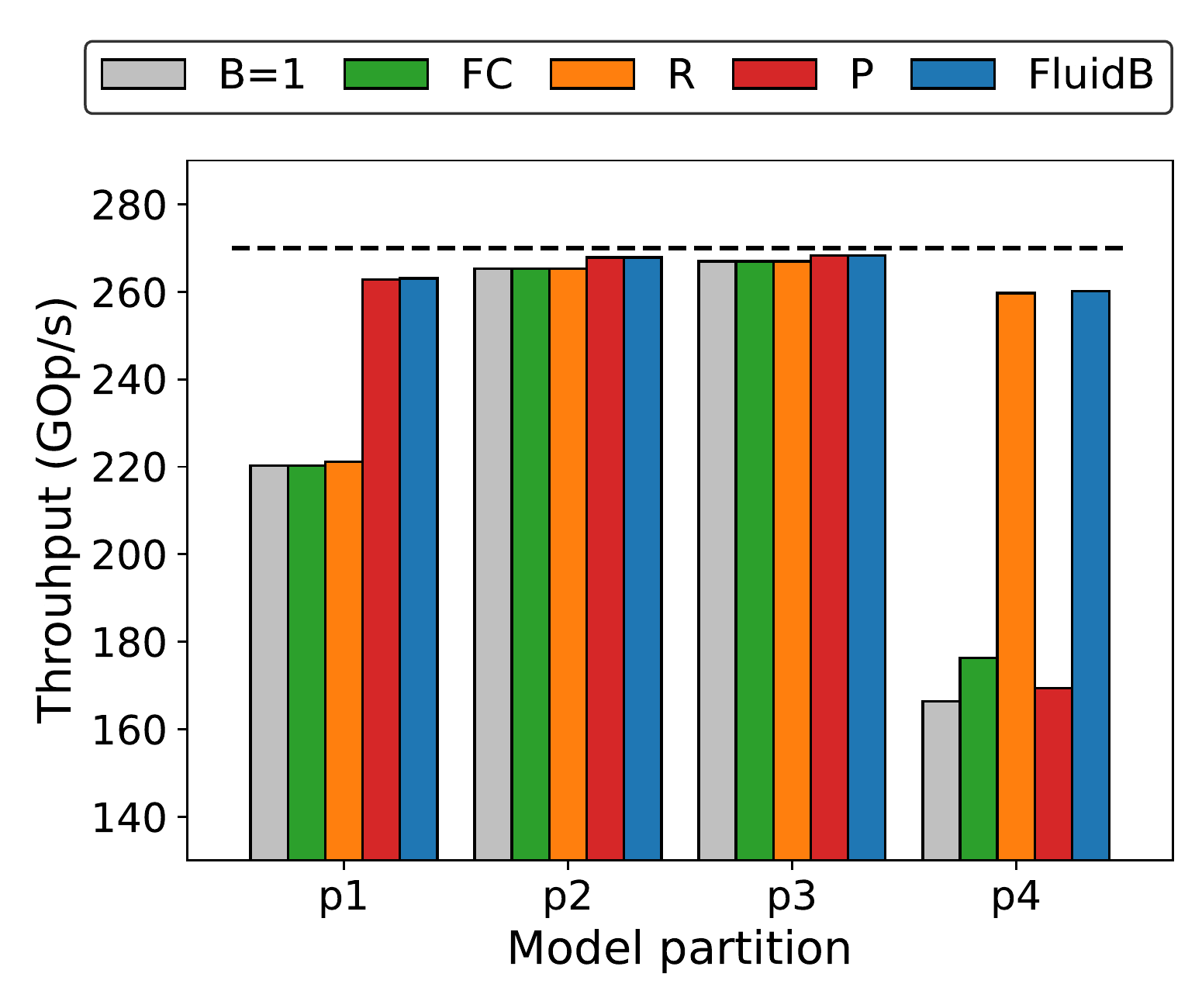}
    \put (-250,6) {\footnotesize (a)}
    \put (-120,6) {\footnotesize (b)}
    \put (-35,81) {\tiny Comp. Roof}
    \vspace{-0.1cm}
    \captionsetup{font=footnotesize	,labelfont=bf}
    \caption{\blue{Different batching strategies applied on a mainstream NPU design~\cite{tpuv4i2021isca} that maps each convolution to a General Matrix Multiply operation, where inputs, weights and output activations are represented by $R$$\times$$P$, $P$$\times$$C$ and $R$$\times$$C$ Toeplitz matrices (see Fig.~\ref{fig:gemm}), respectively, assuming a batch of $B$ samples:}
    \textbf{SERIAL}: $B$$=$$1$ $\vert$ \textbf{FC}: Batching only on FC layers $\vert$ $\mathbf{R/P}$: Appending samples across the row/col dimension of the activation matrices. \textbf{(a)} Impact of early exits (EE) on hardware utilisation with batch processing across two FPGA platforms, targeting a 4-exit ResNet-50 at an arrival rate of 25 samples/s.  Although batching significantly improves the device utilisation for the original network, the stochasticity introduced by early exits radically attenuates this gain. Batches are formed using adaptive batching (AdaptB) (detailed in \S\ref{sec:batch_proc}).   \textbf{(b)} Achieved throughput on ZC706 for ResNet-50 (partitioned in 4 subnets) with different batching strategies adopted by the NPU (B$=$8). Dashed line denotes peak platform performance.
    }
    \vspace{-0.2cm}
    \label{fig:cover}
\end{figure}

However, the inherent dynamicity of smart ecosystems and user behaviour leads to varying inference request rates.
For example, the dynamic adjustment of sampling rates --commonly used to reduce computational load and energy consumption-- affects batch formation at the edge NPU where all samples are streamed. This calls for new hardware solutions that enable the NPU to sustain high performance across variable batch sizes.


At the same time, a growing body of work is developing multi-exit DNNs~\cite{ee_survey2021emdl}. This family of dynamic models attaches intermediate exits throughout the architecture and saves computation by allowing easier samples to exit early. Despite the latency benefits, this mechanism introduces further dynamicity to the batch size, at a subnet level; as samples may exit early at intermediate exits, the batch size shrinks dynamically during inference. As shown in Fig.~\ref{fig:cover}a, this leads to hardware underutilisation for deeper parts of the model and an inefficient use of the NPU, requiring a novel system design approach.




In order to deal with the enhanced dynamicity evident in the deployment of early-exit DNNs in streaming scenarios, based on the above observations we propose a novel SW/HW co-design framework for edge NPU-based serving that employs three core techniques: \textit{i)}~\emph{Fluid Batching}, a hardware-based mechanism that finely adapts the batching strategy 
based on both the instantaneous batch size and the characteristics of each layer, attaining high performance on the NPU across batch sizes (\S\ref{sec:fluidbatch});
\textit{ii)}~an exit- and deadline-aware \emph{preemptive scheduler} that allows the preemption of processing at intermediate exit points and the subsequent merging into larger batches, alleviating the NPU underutilisation due to the reduced batch size in deeper parts of early-exit models (\S\ref{sec:scheduler}); 
\textit{iii)}~\emph{Stackable PEs}, a type of run-time reconfigurable Processing Elements (PEs) that can be dynamically adapted to the dimensions of each layer, counteracting the hardware underutilisation caused by suboptimally mapped operations with minimal overhead (\S\ref{sec:fluidbatch}). 


\section{Background \& Related Work}
\subsection{Batch Processing in Inference Systems}
\label{sec:batch_proc}

Batch processing has been adopted as a key technique towards increasing the inference throughput.
Nonetheless, contrary to the training stage, forming batched inputs during inference is challenging as the processing platform receives DNN inference requests at rates that vary significantly based on the time of day and the number of applications or users that share the same model. Importantly, waiting to form a large-enough batch often has a prohibitive impact on latency.

Existing batching approaches commonly integrate two techniques, \textit{i)}~\textit{model-level} and \textit{ii)}~\textit{adaptive batching}. 
\textit{Model-level batching} dictates that batching is performed at the entire model granularity, \textit{i.e.}~the same batch size $B$ is used for all layers of the DNN. 
\textit{Adaptive batching} (AdaptB), which is employed by modern inference systems~\cite{clipper2017NSDI,mark2019atc,nexus2019sosp,batch2020sc,equinox2021micro} to adapt the batch size to the changing arrival rate, introduces another parameter, the batch-forming timeout window $T_{\text{timeout}}$, resulting in the batching strategy $\left< B_{\text{max}}, T_{\text{timeout}} \right>$.
Concretely, AdaptB dispatches incomplete batches when $T_{\text{timeout}}$ is exceeded, otherwise it issues the platform- or model-dependent maximum batch size $B_{\text{max}}$. 
Closer to our approach, LazyBatching~\cite{lazybatching2021hpca} allows the selective preemption of inference at the layer level. Nonetheless, 
the use of a coarse latency estimator, together with its exit-unaware design, lead to conservative batching decisions, leaving untapped optimisation opportunities. 

\noindent
\textbf{Edge- vs. Cloud-based Inference Systems.} Contrary to their cloud-residing counterparts, edge-based inference systems are constrained in three ways~\cite{edge_ai_hub2022arxiv}: \textit{1)}~edge NPUs have lower resource and energy budget and hence provide lower processing throughput. Thus, in spite of any throughput benefits, the large batch sizes that are common on the cloud (\textit{e.g.}~typically $>$8 and up to 64~\cite{equinox2021micro,lazybatching2021hpca,infaas2021atc}) directly violate the latency service-level objective (SLO) of interactive applications when executing high-accuracy -but costly- models; \textit{2)}~the number of served devices -and in turn the queries per second- is one to two orders of magnitude lower (\textit{e.g.}~$10^2$-$10^3$~queries/s on the cloud vs. tens on the edge~\cite{mlperf_inf2020isca}). Forming large batch sizes would require prohibitive long waiting times to assemble enough samples; \textit{3)}~although lightweight models could be employed to increase throughput, this would degrade the accuracy of the target task and would defeat the purpose of using a server. Fluid Batching alleviates these constraints by maximising the performance even at smaller batch size (see \textit{NPU Ablation} in \S\ref{sec:eval}) and dynamically filling gaps in the active batch as old samples early-exit and new ones arrive. \steve{How does 3 related to edge vs cloud based? } \stelios{In the cloud, we have the compute power to support large batches for large models. In the cloud, we can scale-up the compute, while at the edge, we want better/more efficient utilisation of the hardware.}

\noindent
\blue{\textbf{Batching in Hardware.}}
\blue{From a hardware perspective, existing NPU designs~\cite{equinox2021micro} adopt a \textit{uniform} solution for implementing batching across all layers, which can lead to suboptimal mapping and underutilisation in certain layers. Instead, Fluid Batching (\S\ref{sec:fluidbatch}) introduces a flexible mechanism that adapts the batching strategy on-the-fly at a per-layer basis to maximise resource utilisation and, thus, inference efficiency.}


\subsection{Early-Exit Neural Networks}
\label{sec:dynamic_nns}

Dynamic DNNs come in many shapes and forms \cite{blockdrop2018cvpr,skipnet2018eccv,branchynet2016icpr,fjord2021neurips}. A successful variant of such networks comes in the form of early-exit networks \cite{ee_survey2021emdl}, DNNs with intermediate heads 
where samples can ``exit 
early" based on shallow (and more coarse) features of the original network -- the backbone. This mechanism yields computation savings due to the early termination of inference, while also providing early actionable results during inference 
and allowing for the  progressive refinement of the output prediction. Early-exit networks have successfully been deployed in a multitude of tasks, spanning from image classification~\cite{hapi2020iccad, flexdnn2020sec, frameexit2021cvpr} to semantic segmentation~\cite{mess2022eccv} or even various NLP tasks~\cite{fastbert2020acl,deebert2020acl,patient_bert2020neurips}. On the system side, existing work has so far focused on the hardware-aware design of the exit policy, and the placement and architecture of the exits along the depth of the backbone network~\cite{hapi2020iccad, flexdnn2020sec,mess2022eccv}, the distributed execution of such models across device and server~\cite{distributed_branchynet2017icdcs,spinn2020mobicom,hastening2022tmc} \blue{with \cite{see2020cc} also studying the implications of their deployment in streaming scenarios,} and the co-design of an early-exit model and its hardware accelerator~\cite{edgebert2021micro}. 

All of the above works, however, have focused on single-sample execution and do not consider the hardware utilisation impact of such a decision at inference time. In contrast, we embrace larger batch sizes in the streaming setting and investigate an efficient way of scheduling such stochastic workloads on NPUs under tight latency budgets.



\subsection{NPU Architecture Design}
\label{sec:npus}
\begin{figure}[t]
    \centering
    \includegraphics[width=0.8\columnwidth,trim={0cm 3.75cm 14.5cm 0cm},clip]{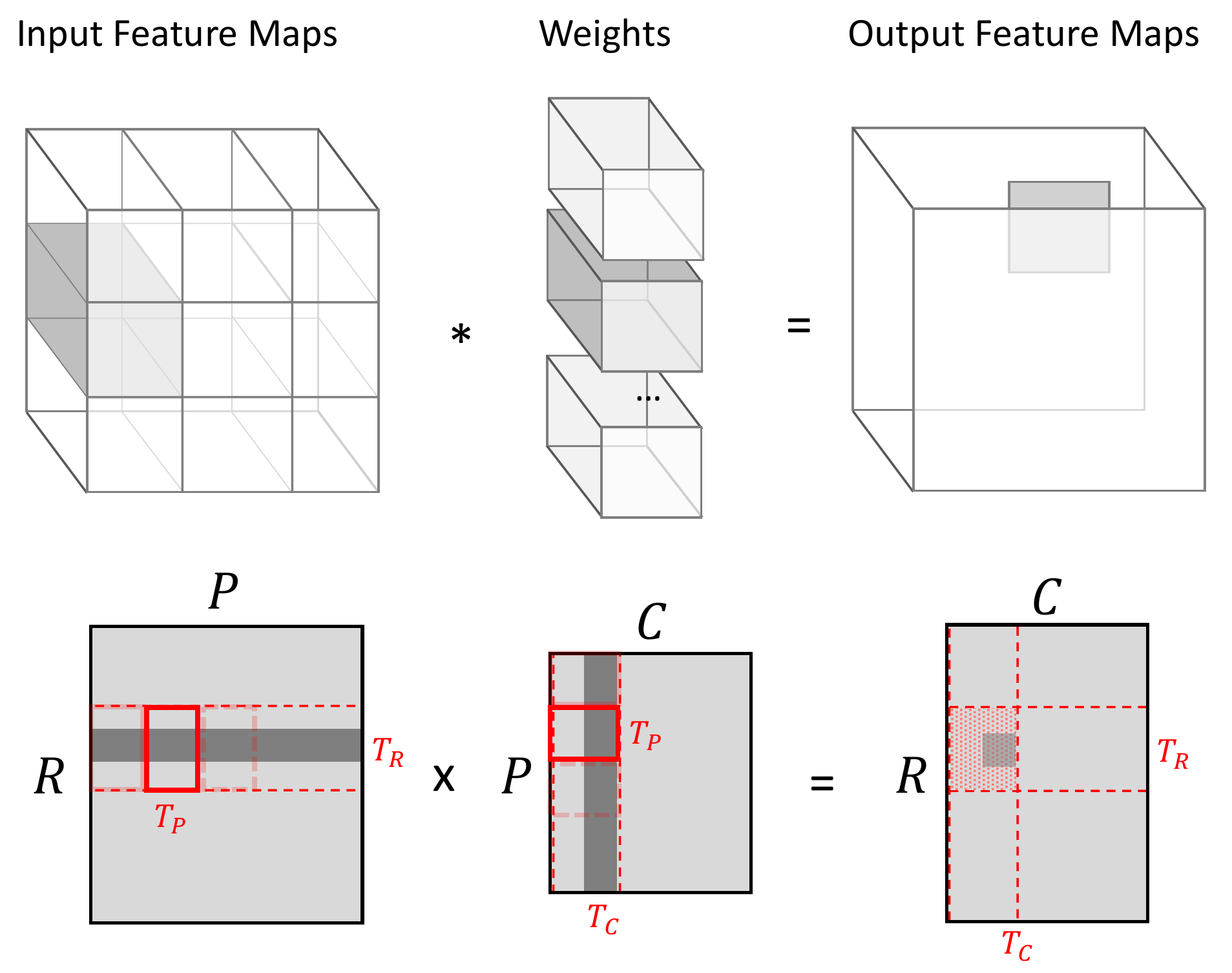}
    \captionsetup{font=small,labelfont=bf}
    \caption{The computation of a convolutional layer, mapped as GEMM.}
    \vspace{-0.1cm}
    \label{fig:gemm}
\end{figure}

General matrix multiply (GEMM) comprises the most widely optimised computational kernel that lies at the heart of most NPU accelerators due to its ability to support both the convolutional (CONV) and fully-connected (FC) layers~\cite{tpuv4i2021isca} of CNNs, as well as the various matrix multiplications of Transformer models~\cite{sigma2020hpca}. This paper considers a generic NPU design~\cite{shidiannao2015isca,cascadecnn2018,unzipfpga2021fccm} that represents a large potion of actual deployed NPUs, where each layer's input feature maps, weights and output feature maps are formed into $R$$\times$$P$, $P$$\times$$C$ and $R$$\times$$C$ Toeplitz matrices respectively, that are stored in the off-chip memory. The execution of each layer is then reduced to a tiled GEMM operation, as depicted in Fig.~\ref{fig:gemm}. Typically, tile sizes across each dimension ($T_R, T_P, T_C$) are tightly coupled with architectural characteristics of the accelerator (concerning both the computation element and on-chip memory structure) and tuned based on the target workload through design space exploration (DSE). In this work, the adopted NPU consists of $T_C$ processing elements (PEs), each comprising a Multiply-Accumulate (MAC) tree of width $T_P$ (Fig.~\ref{fig:npu}b). Parameter $T_R$ controls the pipeline depth, while on-chip memory buffers matching the input, weight and output tile dimensionalities and allowing for double buffering are instantiated.

Under this shared accelerator design, it is common that not all layers of the target model can be efficiently mapped, due to a mismatch between layer dimensions and the accelerator's configuration. As a result, the resources of the NPU are not fully utilised throughout the inference process, leading to suboptimal performance. To remedy this inefficiency, in this paper we propose Stackable PEs (\S\ref{sec:fluidbatch}), a dynamically adjustable PE structure that allows the on-the-fly re-purposing of the underutilised resources in order to exploit different parallelism dimensions with minimal hardware overhead. 

\section{Methodology}



Fig.~\ref{fig:overview} depicts an overview of the proposed system. Inference requests associated with a common model are coming into a \textit{queue}~$\bigl(\circled{1}\bigr)$ to be processed by a multi-exit model resident on the \textit{NPU}~(\S\ref{sec:fluidbatch}). 
The \textit{scheduler} (§\ref{sec:scheduler}), running on CPU, reads from the queue, forms a batch into a \textit{buffer} and submits inference jobs to the \textit{NPU}~$\bigl(\circled{2}\bigr)$. Upon meeting a preemption point, \textit{i.e.}~an intermediate classifier where some of the samples are expected to exit early, an early-exit event is triggered~$\bigl(\circled{3}\bigr)$ and the number of exiting samples, $B_{\text{exit}}$, is communicated to the scheduler~$\bigl(\circled{4}\bigr)$. At this point, the scheduler \textit{selectively} preempts execution~$\bigl(\circled{5}\bigr)$ and a new batch of size $B_\text{incr}$ is formed and dispatched to run until the previous preemption point~$\bigl(\circled{6}\bigr)$. Subsequently, the halted and the new batch are merged and execution is resumed, benefiting from the higher efficiency of the increased batch size, until the next exit is met. The whole process is orchestrated by the scheduler and the \textit{Fluid Batching Engine}~$\bigl(\circled{7}\bigr)$ which tailor the adaptable batching and processing components of the NPU~$\bigl(\circled{8}\bigr)$ based on the current batch size, the model at hand, the latency SLO and the arrival rate of the incoming samples, aiming to maximise efficiency. Next, we delve into the details of each component.

\subsection{Edge NPU Design with Fluid Batching}
\label{sec:fluidbatch}





\begin{figure*}[t]
    \centering
    \includegraphics[width=1.9\columnwidth,trim={14cm 5mm 0cm 0cm},clip]{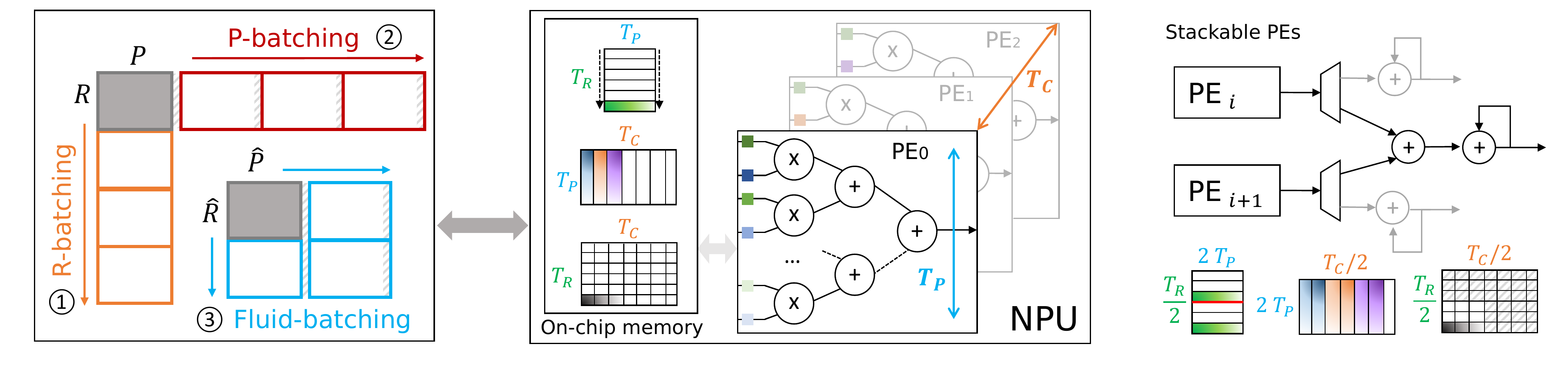}
    \put (-490,7) {\footnotesize (a)}
    \put (-337,7) {\footnotesize (b)}
    \put (-135,7) {\footnotesize (c)}
    \captionsetup{font=small,labelfont=bf}
    \caption{\textbf{(a)} Input matrix formation with different batching strategies ($B$$=$$4$), \textbf{(b)} NPU architecture, \textbf{(c)} Stackable PEs ($k$$=$$2$). }
    \vspace{-0.1cm}
    \label{fig:npu}
\end{figure*}

Under the GEMM formulation (\S\ref{sec:npus}), existing systems typically realise batching of $B$ samples by appending activation matrices uniformly along the \textit{row} dimension for all layers (see Fig.~\ref{fig:npu}a~$\bigl(\wcircled{1}\bigr)$), so that $\hat{R}$$=$$R$$\cdot$$B$, allowing for better reuse of the weight matrix. This has been proven particularly effective in counteracting the low computation-to-communication ratio of the memory-bound fully-connected layers, where $R$$=$$1$ when batching is not used~\cite{cascadecnn2018}. Adopting the same approach in convolutional layers, across either the $R$~$\bigl(\wcircled{1}\bigr)$ or $P$~$\bigl(\wcircled{2}\bigr)$ dimension to facilitate further parallelism between the samples of a batch, has also shown considerable performance gains~\cite{equinox2021micro, hadjis2015caffe}.

Nonetheless, in all cases, batching and parallelism are \textit{statically} applied across all layers, whereas NPUs are typically optimised for a \textit{fixed} batch size. Fig.~\ref{fig:cover}b reports the achieved throughput of different static batching strategies for a ResNet-50 backbone on the examined NPU, with its layers partitioned into groups of equivalent workload. Evidently, different parts of the model benefit the most from different batching implementations, as shallow and deep layers demonstrate fundamental variability in matrix dimensionalities, leading to distinct mapping inefficiencies.

\vspace{1mm}
\noindent
\textbf{Flexible Batch Processing Mechanism.}
In order to remedy the inefficient mapping of static batching approaches, Fluid Batching generalises existing strategies by dynamically selecting the breakdown of samples that are appended in each matrix dimension ($R$, $P$), based on the running layer $l$ and the active batch size $B_{\text{act}}$. As such the NPU is able to exploit parallelism across different samples of a batch (previously being pipelined), in cases where other parallelism dimensions lead to an inefficient mapping on the available resources. In practice, Fluid Batching is realised by modifying the Direct Memory Access (DMA) to the off-chip memory, to affect the Toeplitz matrix formation process for each batch.
Formally:
\begin{equation}
  \small
  \begin{split}
        \hat{R}^{(l)}=&B_R^{(l,B_{\text{act}})} \cdot R^{(l)} \quad \text{and} \\  \hat{P}^{(l)}=&(B_{\text{act}} - B_R^{(l,B_{\text{act}})} +1) \cdot (P^{(l)} + P^{(l)} \% T_P )
\end{split}
\end{equation}
where $B_R$$=$$f(l,B_{\text{act}}) \in [1,B_{\text{max}}]$ is provided by the Fluid Batching engine (described below) that controls the batching strategy for each layer at run time through a fine-grained mechanism, aiming to improve efficiency by eliminating resource underutilisation; and $\%$ denotes the modulo operation, used to add zero ``guard'' elements across the $P$ dimension (Fig.~\ref{fig:npu}a) to prevent interference between different samples of the batch. As also illustrated in Fig.~\ref{fig:npu}a, R- and P-batching form special cases of Fluid Batching~$\bigl(\wcircled{3}\bigr)$ for $B_R$$=$$B_{\text{act}}$ and $1$, respectively, along with other hybrid schemes. Notably, through DMA handling, the result of Fluid Batching's computation remains a $B_{\text{act}} \cdot R \times C$ matrix, as in the case of R-batching, and facilitates the adoption of a different batching scheme at the subsequent layers.

\begin{figure}
    \centering
    \includegraphics[width=0.9\columnwidth,trim={1cm 3.5cm 14cm 4cm},clip]{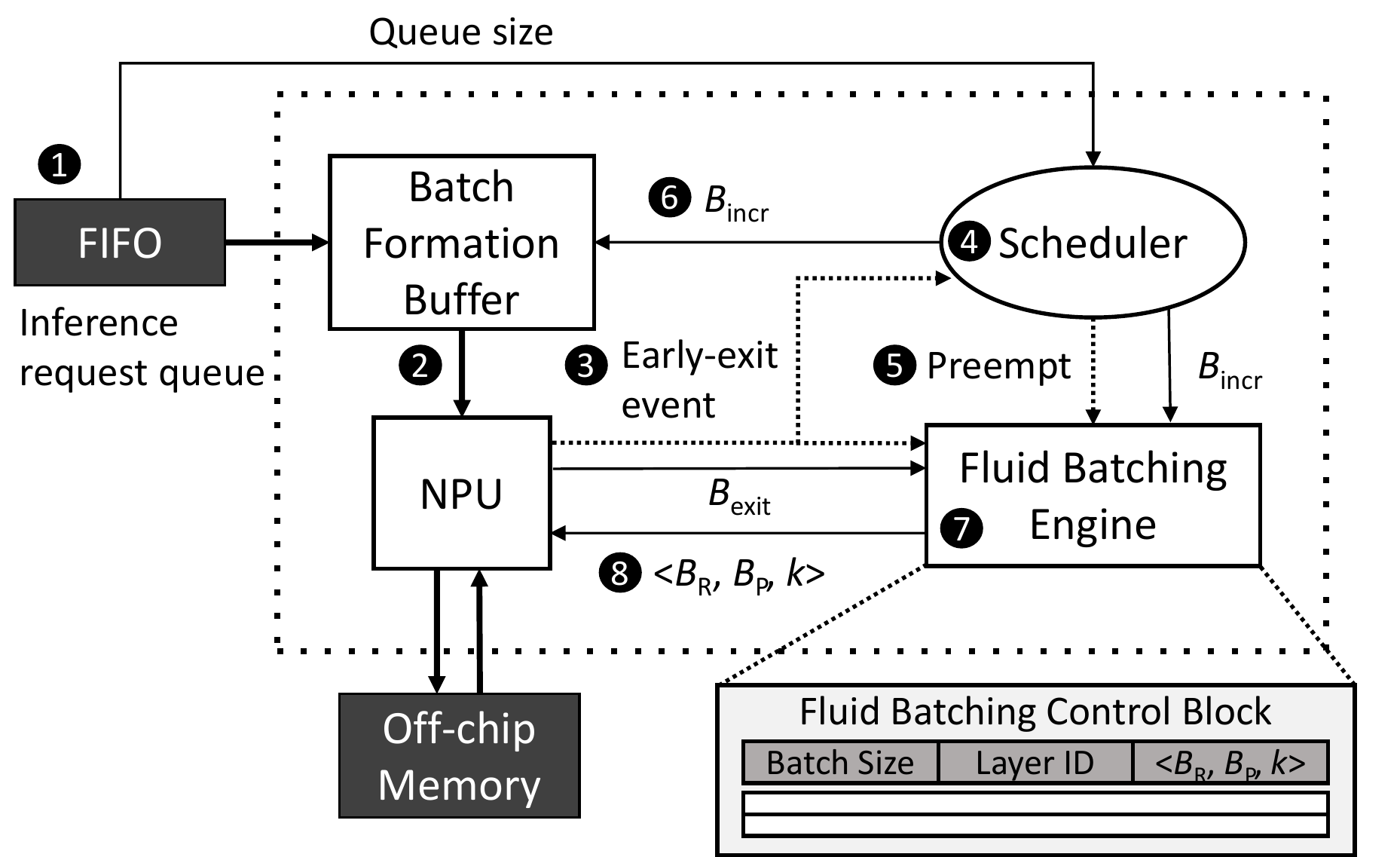}
    \captionsetup{font=footnotesize	,labelfont=bf}
    \caption{Overview of the proposed system.}
    \vspace{-0.3cm}
    \label{fig:overview}
\end{figure}

\vspace{1mm}
\noindent
\textbf{Stackable Processing Elements.} 
Since all $L$ layers {\small $\left<R^{(l)},P^{(l)},C^{(l)}\right>$} of a model are mapped to the same NPU {\small $\left<T_R,T_P,T_C\right>$}, naturally some compute or processing elements will remain underutilised during the execution of layers with deviating feature volume shape (where the globally optimised tile sizes exceed or cannot perfectly divide the corresponding matrix dimension). The proposed batching strategy is able to remedy the inefficient mapping of CONV and FC layers to hardware, by taking advantage of the increased dimensionality offered by batching in a flexible manner. In essence, Fluid Batching configures the input matrices so that different samples can be processed concurrently by the NPU, in layers where some processing elements remain underutilised, \textit{e.g.}~when $C^{(l)}$$<$$T_C$. Expectedly, this flexibility is decreased when the batch size remains small, which can often occur during low-traffic periods in streaming scenarios. Additionally, in some layers, mapping inefficiencies caused by other factors can remain after this optimisation is applied, \textit{e.g.}~when $P^{(l)}$$<$$T_P$, leading to a persistent device underutilisation. 

To further improve the inference efficiency across the spectrum, in this section we introduce a mechanism, termed Stackable PEs, that empowers the NPU with further flexibility on the allocation of its resources. Stackable PEs allow partial on-the-fly restructuring of the NPU's processing elements, by re-purposing their MAC units and on-chip memory buffers to better fit the shape of the input and weight matrices. Through the design depicted in Fig.~\ref{fig:npu}c, a {\small $\left<T_R,T_P,T_C\right>$} NPU can be transformed on-the-fly to an alternative {$\left< \lfloor T_R/2\rfloor,k\cdot T_P, \lfloor T_C/k\rfloor \right>$} design, where $k$$\in$$\{\rfrac{1}{2},2\}$, with minimum hardware overhead. This approach is essentially trading coarse- to fine-grained parallelism, while ensuring adequacy of the instantiated memory buffers, in order to maintain as many PEs as possible active during the execution of each layer. The most efficient PE configuration for each layer of the target workload is identified during DSE and controlled by the Fluid Batching Engine during inference through its $k^{(l)}$ signal. 

\vspace{1mm}
\noindent
\textbf{Design Space Exploration (DSE).}
In this context, we enhance the conventional DSE mechanism for edge NPUs in a two-fold manner. First, we consider the attainable performance of each candidate design point \textit{across different batch sizes}, effectively co-optimising the hardware architecture for various batching scenarios. Second, for each visited NPU design point $d$$=$$\left<T_R,T_P,T_C\right>$, we expose the design spaces of Fluid Batching and Stackable PEs to the optimisation, leading to {\small $\left<d, \boldsymbol{k}, \boldsymbol{B}_R\right>$}. Parameters {\small $\boldsymbol{k}$$\in$$\{1/2, 1, 2 \}^{L\times B_{\text{max}}}$} ($k$$=$$1$ is the default PE configuration) and {\small $\boldsymbol{B}_R$$\in$$[1, B_{\text{max}}]^{L\times B_{\text{max}}}$} hold the Stackable PEs and Fluid Batching configuration for each layer and batch size, and are used to populate the Fluid Batching Engine (Fig.~\ref{fig:overview}). Hence, we cast DSE as a formal optimisation problem:
\vspace{-0.5mm}
\begin{equation}
    \small
    \vspace{-0.5mm}
    \resizebox{0.85\linewidth}{!}{
        $\left<d^*, \boldsymbol{k}, \boldsymbol{B}_R \right> =  \underset{d}{\arg\max}  \underbrace{\sum_{b=1}^{B_{\text{max}}} w_b}_{\text{for each batch size}} \underset{\boldsymbol{k}, \boldsymbol{B}_R} \max \quad T \Bigl(\left<d, \boldsymbol{k}, \boldsymbol{B}_R\right>, W, b \Bigr)$
        }
\end{equation}
where weight $w_b$ is used to control the contribution of each batch size. We estimate the performance, $T(\cdot)$, in GOp/s of each examined design point on a given workload $W$ with a combination of analytical and roofline modelling~\cite{cascadecnn2018}. Finally, the highest performing design is obtained through exhaustive search.

\begin{figure}[t]
    \centering
    {
    \includegraphics[width=0.95\columnwidth,trim={4.2cm 6.75cm 12cm 5.5cm},clip]{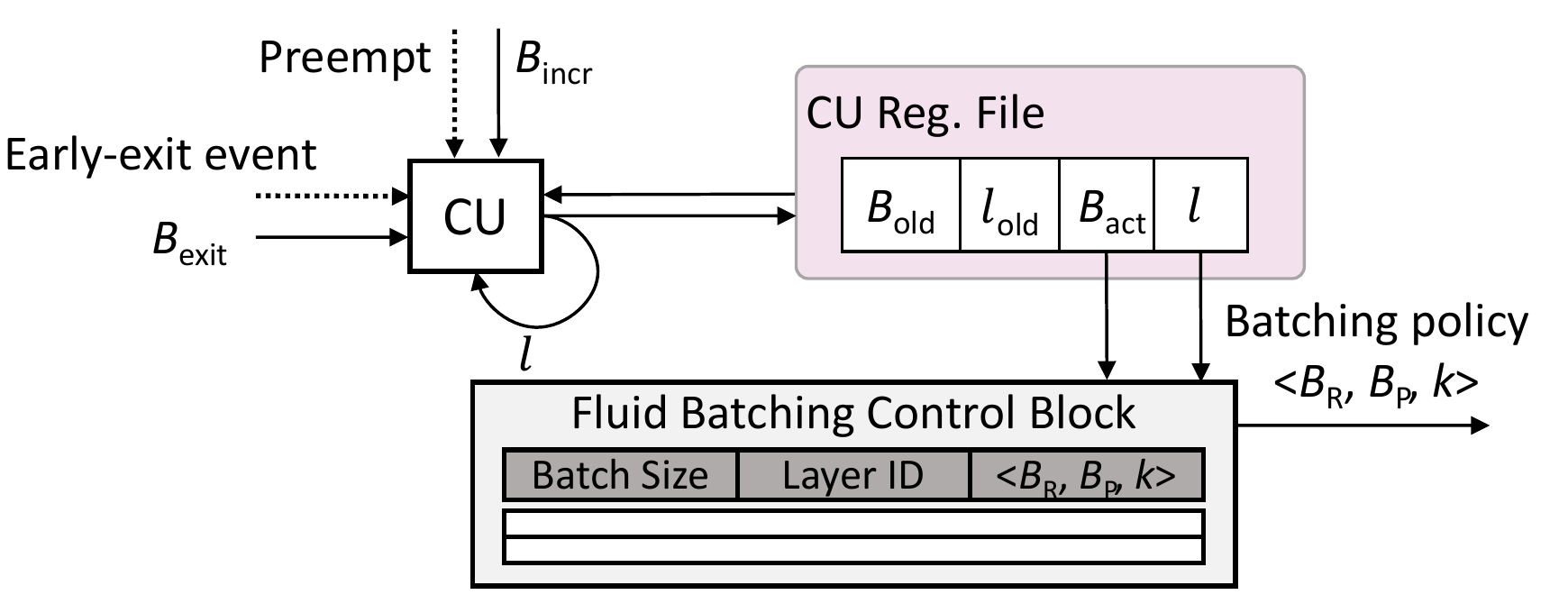}
    }
    \vspace{1mm}
    \captionsetup{font=footnotesize,labelfont=bf}
    \caption{Microarchitecture of the Fluid Batching Engine.}
    \vspace{-2mm}
    \label{fig:fbe_arch}
\end{figure}

\vspace{1mm}
\noindent
\textbf{Microarchitecture.} 
Fig.~\ref{fig:fbe_arch} shows the hardware design of the Fluid Batching Engine, comprising the Fluid Batching Control Block (FBCB) and a Control Unit (CU). The FBCB constitutes a look-up table that stores the highest performing batching policy and Stackable PEs configuration $\left<B_R, B_P, k\right>$ for each layer of the given DNN and for different batch sizes. As such, the FBCB contains $L$$\times$$B_{\text{max}}$ entries. We store only $k$ and $B_R$, and derive $B_P$ as $B_{\text{act}}$$-$$B_P$. With $B_R$ bounded by $B_{\text{max}}$, we represent each layer's $B_R$ entry with {\small$\lceil \log_2(B_{\text{max}}) \rceil$} bits and encode $k$'s three states with 2 bits.

The CU uses $B_{\text{exit}}$ and  $B_{\text{incr}}$ (\S\ref{sec:scheduler}) to keep track of the active batch size ($B_{\text{act}}$). When a new layer is to be processed, the CU uses $B_{\text{act}}$ and the layer index $l$ to address the FBCB and fetch the correct batching policy. Finally, the batching policy is used to configure the NPU's DMA controllers. This affects how the NPU reads and writes the input and output activation matrices from and to the off-chip memory.

\subsection{Exit-Aware Preemptive Scheduler for Early-Exit DNNs}
\label{sec:scheduler}

Conventional inference systems~\cite{clipper2017NSDI,mark2019atc,nexus2019sosp,batch2020sc,equinox2021micro} utilise the same batch size for the whole DNN. In particular, once a batch of inputs has been dispatched to the accelerator, future arriving samples have to wait until the processing of the whole batch has completed. The limitations of this approach become especially evident when processing early-exit DNNs, where the active batch size can change dynamically through the depth of the network. In this case, the status-quo model-level batching constrains the DNN to execute until the end with a reduced batch size, even if it severely underutilises the accelerator's resources.

We propose an exit-aware preemptive scheduler that considers both the active batch size and the SLO to balance latency, throughput and SLO satisfaction. In contrast to existing systems, 
we introduce a scheduling granularity at the subnet level, with boundaries at the intermediate exits. 
As such, the already running \textit{active} batch can be preempted and new samples can be processed in an interleaved manner. Concretely, the scheduler selectively preempts execution at the exit points and dispatches a new batch so that it can catch up. Next, the new samples are merged with the preempted samples to form a larger batch and resume execution with increased hardware utilisation.
Formally, we parametrise this policy as {\small $\left< B_{\text{max}}, T_{\text{SLO}} \right>$}, where {\small $T_{\text{SLO}}$} is the latency SLO. 
Our scheduler launches execution as soon as the first sample enters the request queue (Fig.~\ref{fig:overview}). Algorithm~\ref{alg:scheduling} details our scheduling method. Next, we point to the related lines as we describe our method.


\vspace{1mm}
\noindent
\textbf{Preemptible Points.} 
Given an $L$-layer DNN with a set of early exits $\mathcal{E}$, our scheduler allows for preemption only at the early-exit points $i$$\in$$\mathcal{E}$. This design decision is based on two key insights. First, preemption is beneficial only when there are dynamic changes in the batch size and hence evaluating the preemption criterion elsewhere leads to redundant computation. 
Second, treating every layer as preemptible would introduce prohibitively high overhead~\cite{lazybatching2021hpca}; the scheduler would be invoked too regularly, inducing excessive computations and interruption of the NPU's operation; \textit{e.g.}~our approach yields 16.6$\times$ fewer scheduler invocations for a 4-exit ResNet-50 compared to LazyBatching's layerwise approach~\cite{lazybatching2021hpca}.

\begin{figure}[t]
    \centering
    {
    \includegraphics[width=.99\columnwidth,trim={7.5cm 5cm 6.25cm 3cm},clip]{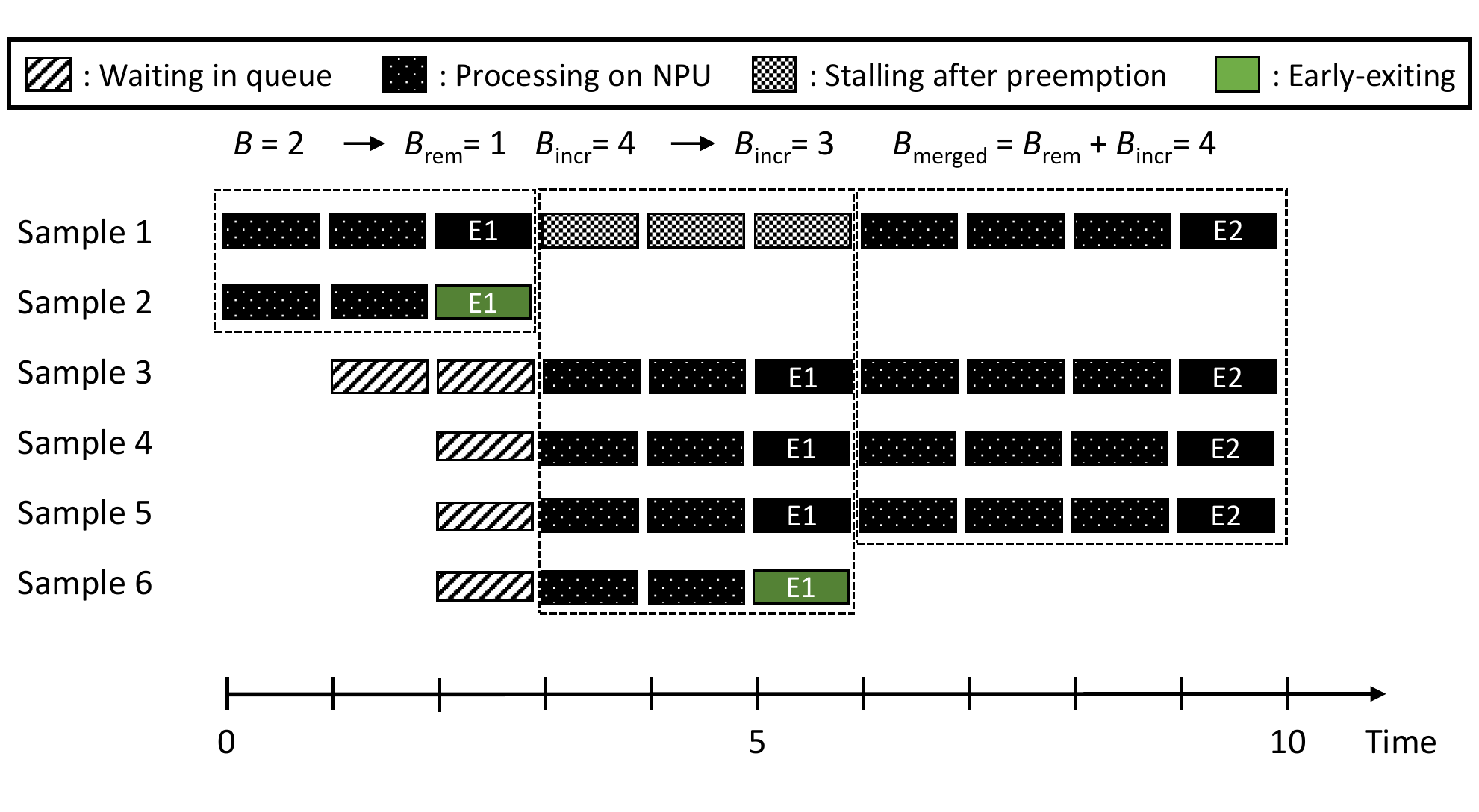}
    }
    \vspace{-5mm}
    \captionsetup{font=footnotesize,labelfont=bf}
    \caption{Scheduling timeline for a 2-exit (E1, E2) DNN with $B_{\text{max}}$$=$$8$. After reaching E1, sample 2 exits early. The scheduler evaluates the SLO-aware preemption criterion and issues a preemption with $B_\text{incr}$$=$$3$. Sample 1 is preempted and samples 3 to 6 are processed up to E1. Sample 6 exits early. The scheduler determines that further preemptions would introduce SLO violations and merges the remaining samples, leading to a batch size of 4 up to exit E2.}
    \label{fig:sched_diagram}
    \vspace{4mm}
\end{figure}

\vspace{1mm}
\noindent
\textbf{Preemption Mechanism.}
Fig.~\ref{fig:sched_diagram} illustrates the operation of our scheduler.
Upon preemption at exit $i$, the intermediate results of the \textit{remaining} active batch of size $B_{\text{rem}}$$=$$B_{\text{act}}$$-$$B_{\text{exit}}$ are written back to the off-chip memory (line~7) and a new batch of samples is issued (lines~\mbox{9-11}). The new batch size is determined as {\small $B_{\text{incr}}$$=$$\min (N_Q, B_{\text{slack}})$}, where {\small \mbox{$B_{\text{slack}}$$=$$B_{\text{max}}$$-$$B_{\text{rem}}$}} 
and {\small $N_Q$} is the instantaneous queue size. When the new samples reach the preemption point (line~14), $B_{\text{incr}}$ might have been reduced as some samples might have exited at the earlier exits. 
Batch backfilling is accomplished per exit and non-recursively, \textit{i.e.}~no nested preemption for intermediate exits. This allows us to have bounded stalling time for the preempted samples while maximising hardware utilisation. 
Finally, the rest of the DNN is executed with a merged batch size of {\small $B_{\text{merged}}$$=$$B_{\text{rem}}$$+$$B_{\text{incr}}$} (line~17), until the next preemption point (line~6). 



\vspace{1mm}
\noindent
\textbf{Preemption Criterion.}
We introduce an SLO-aware preemption criterion that aims to minimise SLO violations. As a first step, the scheduler estimates the remaining time $T_{\text{slack}}$ until the latency SLO is reached for the oldest sample in the active batch:
\begin{equation}
    \footnotesize
    T_{\text{slack}} = T_{\text{SLO}} - \left( T_{\text{wait}} + T_{\text{exec}}^{\text{so-far}} \right)
    \label{eq:slo_slack}
\end{equation}
where $T_{\text{wait}}$ is the queue waiting time of the oldest sample in the active batch and  $T_{\text{exec}}^{\text{so-far}}$ is the execution time passed until the preemption point was reached. The final criterion (line~12) is:
%
\begin{equation}
    \footnotesize
    T_{\text{overhead}} < T_{\text{slack}} \quad \text{with} \quad T_{\text{overhead}} = T_{0:i}^{B_{\text{incr}}} + T_{i+1:L-1}^{B_{\text{merged}}}
    \label{eq:preempt_criterion}
\end{equation}
where $T_{i:j}^{B}$ is the execution time from exit $i$ to $j$ (inclusive) with batch size $B$,
capturing the time for the \textit{new} batch to reach the $i$-th exit and for the \textit{merged} batch to complete the inference. As $B_{\text{incr}}$ might become smaller due to its own early-exiting samples, the actual latency can be smaller and hence this approach can lead to a slight overestimation of the overhead. Nonetheless, it constitutes an upper bound and hence prevents the scheduler from introducing additional SLO violations. 

%

\noindent
\textbf{Exit-Level Latency Prediction.}
Leveraging the deterministic dataflows of modern NPUs, we developed a performance model to estimate the per-exit latency. Alternatively, if the NPU architecture is unknown, we can profile the average per-exit latency.
At design time, we measure the per-exit latency by varying the batch size from 1 to $B_{\text{max}}$ and we bookkeep the results. Upon deployment, the scheduler loads the results in a {\small $N_{\text{exits}}$$\times$$B_{\text{max}}$} look-up table and uses it at run time to evaluate the preemption criterion and guide preemption decisions.


\noindent
\textbf{Overhead.}
As we are targeting a streaming setting, our scheduler picks $B_\text{incr}$ samples from the top of the queue when forming new batches, so the scheduling complexity is $O(1)$. Samples can terminate out-of-order, with the reordering happening afterwards using the sample ID. The required input/output buffers in the off-chip memory are allocated upon initialisation to be large enough to accommodate the largest activation matrix for $B_{\text{max}}$, hence avoiding run-time memory management and the associated latency. Finally, as we preempt active batches at the end of an exit's last layer's execution, the output activations are already stored in the off-chip memory, avoiding the need for explicit checkpointing operations. The main overhead is that the memory transfer of the new input samples is not overlapped with computation. However, our empirical evaluations showed that this contributed at most 0.05\% in latency across instances.

\SetArgSty{textnormal} 
\setlength{\textfloatsep}{0pt}
\begin{algorithm}[!t]	
	\footnotesize
	\SetAlgoLined
	\LinesNumbered
	\DontPrintSemicolon
	
        \KwIn{Model $m$ with exits $\mathcal{E}$}
        \nonl
	\myinput{Latency SLO $T_{\text{SLO}}$}
	\nonl
        \myinput{Maximum batch size $B_{\text{max}}$}
        

        \nonl
        \;
        \textit{/* --- Initialise batch size and exit index --- */}
        
        $B_{\text{act}} \leftarrow \min(N_Q, B_{\text{max}})$
        
        $i \leftarrow 1$

        \nonl
        \;
        \textit{/* --- Process until the end of the network --- */}
        
        \While(\Comment{Until having reached the last exit}){$i < |\mathcal{E}|$}{
            ProcUntilExit($m, B_{\text{act}}, i$) \hfill\Comment{Process up to the $i$-th exit}
    
            $B_{\text{rem}} \leftarrow B_{\text{act}} - B_{\text{exit}}$

            $B_{\text{act}} \leftarrow B_{\text{rem}}$

            \nonl
            \;
            \While(\Comment{If there is room in the batch}){$B_{\text{rem}} < B_{\text{max}}$}{
                $B_{\text{slack}} \leftarrow B_{\text{max}} - B_{\text{rem}}$
    
                $B_{\text{incr}} \leftarrow \min(N_Q, B_{\text{slack}})$

                \nonl
                \;
                \uIf(\Comment{SLO-aware preemption}){$T_{\text{overhead}} < T_{\text{slack}}$}{
                    PreemptActiveBatch($\cdot$)
    
                    ProcFromStartUntilExit($m, B_{\text{incr}}, i$)
    
                    $B_{\text{merged}} \leftarrow B_{\text{rem}} + B_{\text{incr}} - B_{\text{exit}}$
                    
                    \nonl
                    \;
                    $B_{\text{rem}} \leftarrow B_{\text{merged}}$

                    $B_{\text{act}} \leftarrow B_{\text{merged}}$
                }
                \Else{
                    break
                }
            }
            $i \leftarrow i + 1$ \hfill\Comment{Move to next exit}
        }
    

	\caption{\footnotesize 
	Exit-Aware Preemptive Scheduling Method
	}
	\label{alg:scheduling}	
\end{algorithm}

\begin{figure*}[t]
\vspace{-0.2cm}
\centering
\includegraphics[width=0.85\textwidth,trim={0cm 12cm 0cm 0cm},clip]{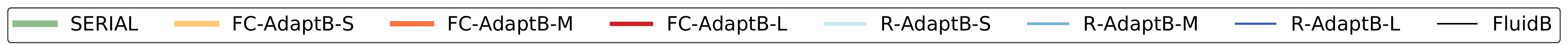}
\vspace{-8mm}
\end{figure*}
\begin{figure}[t]
    \vspace{-0cm}
    \centering
    \subfigure[Processing Rate vs Arrival Rate.]{\includegraphics[width=0.22\textwidth,trim={0cm 0cm 0cm 4.04cm},clip]{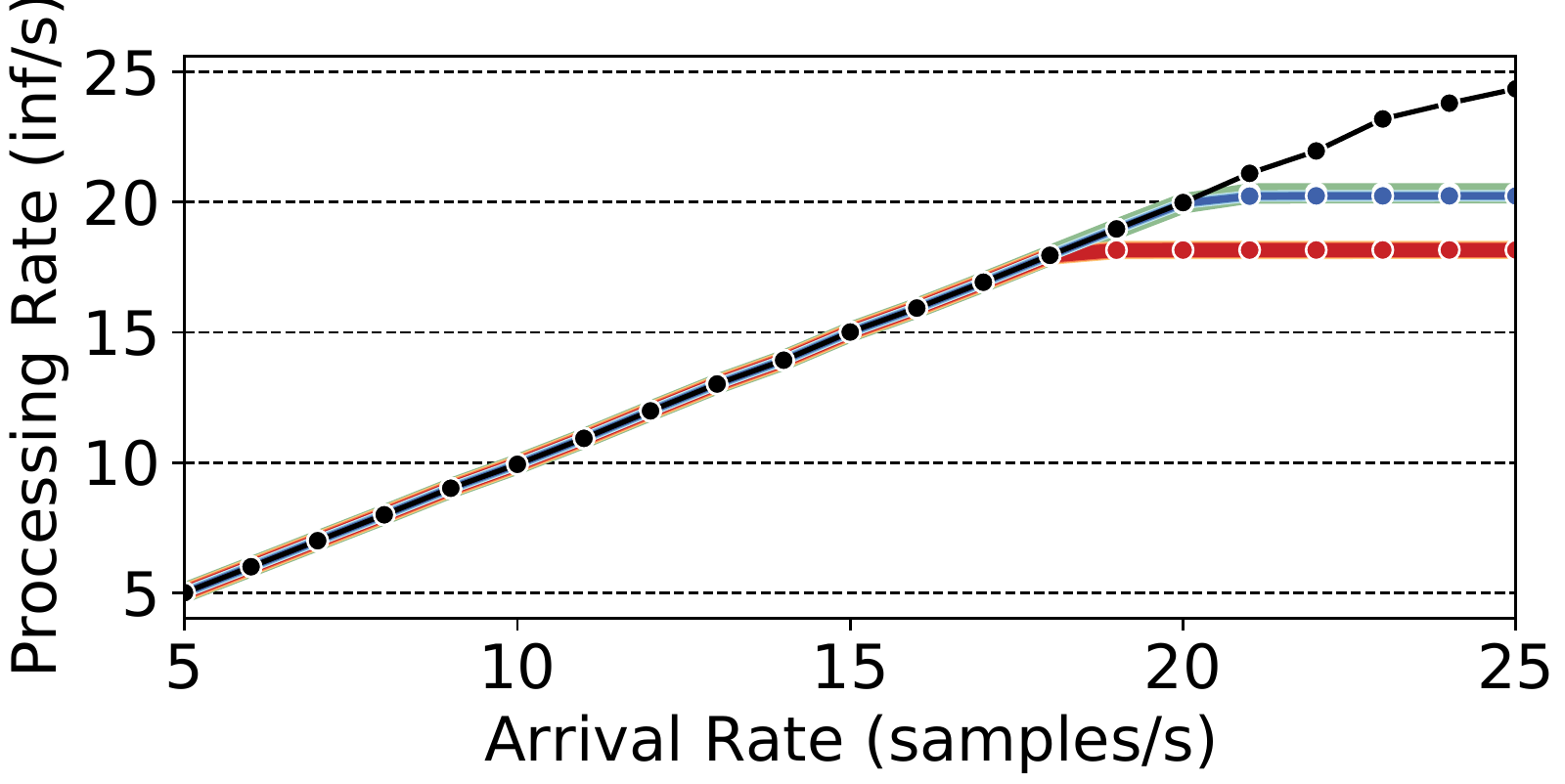}} 
    \subfigure[Avg. Latency vs Arrival Rate.]{\includegraphics[width=0.22\textwidth,trim={0cm 0cm 0cm 4.04cm},clip]{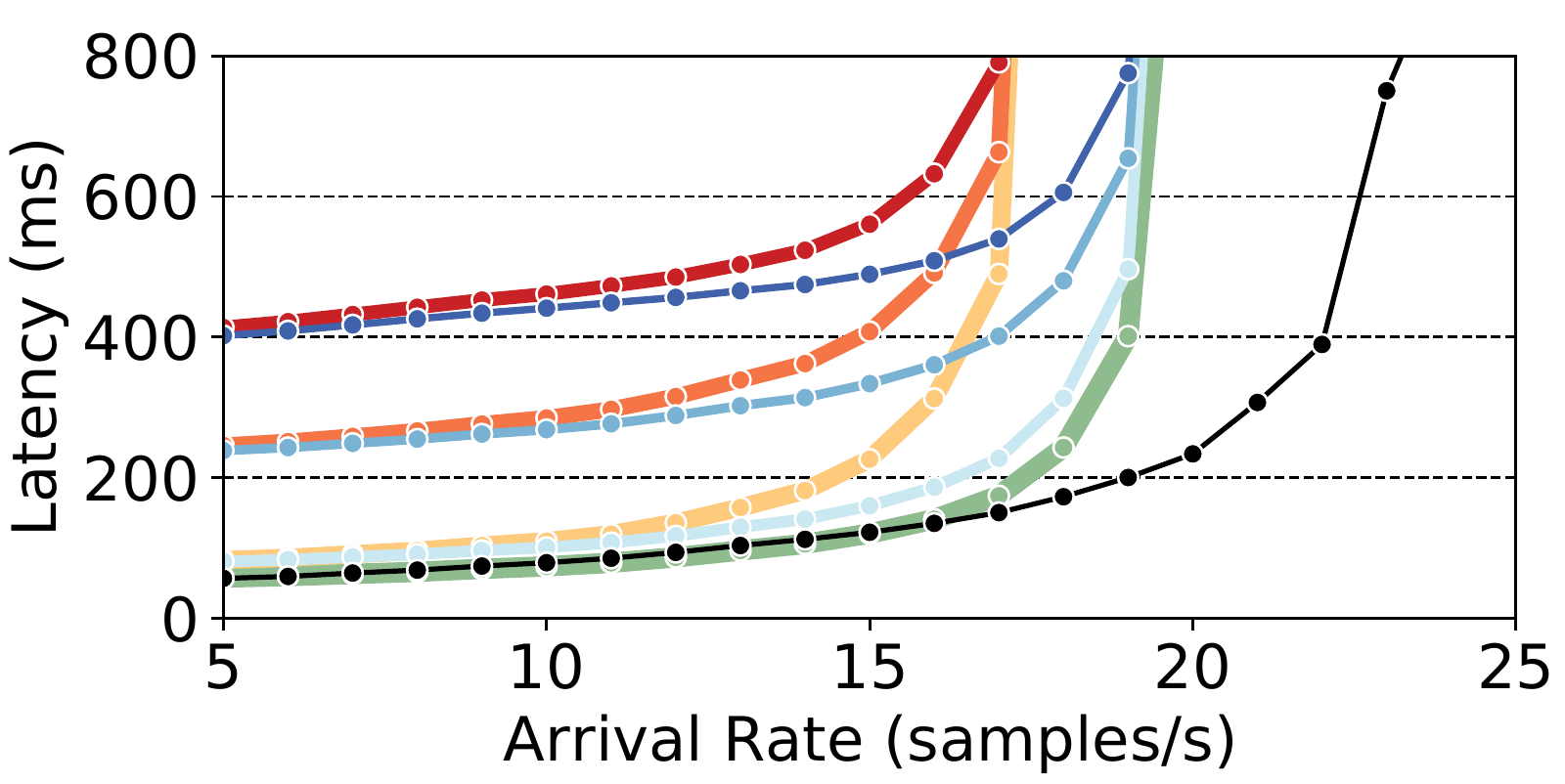}} 
    \subfigure[Utilisation vs Arrival Rate.]{\includegraphics[width=0.22\textwidth,trim={0cm 0cm 0cm 4.25cm},clip]{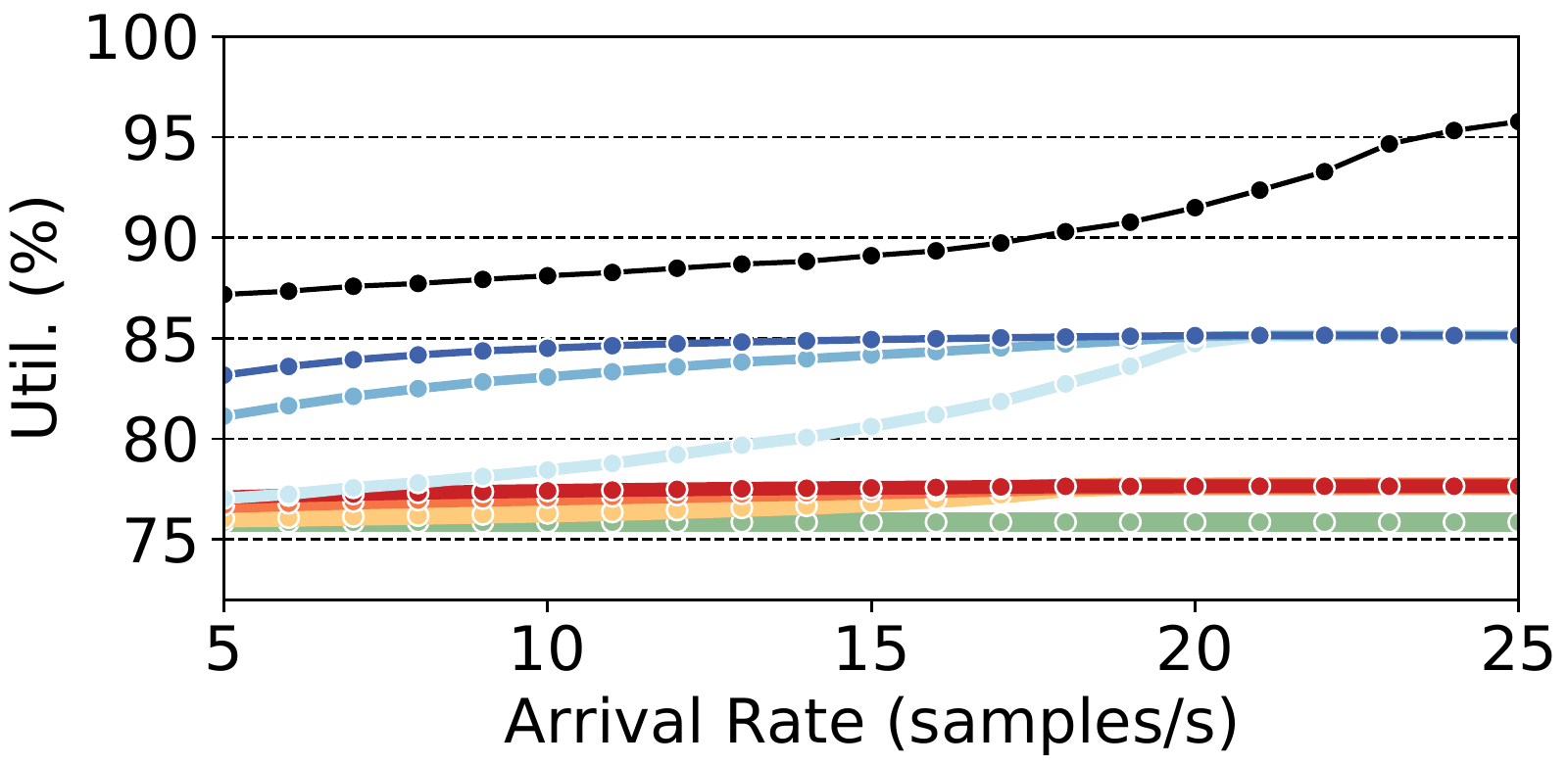}}
    \subfigure[Tail Latency vs Arrival Rate.]{\includegraphics[width=0.22\textwidth,trim={0cm 0cm 0cm 4.25cm},clip]{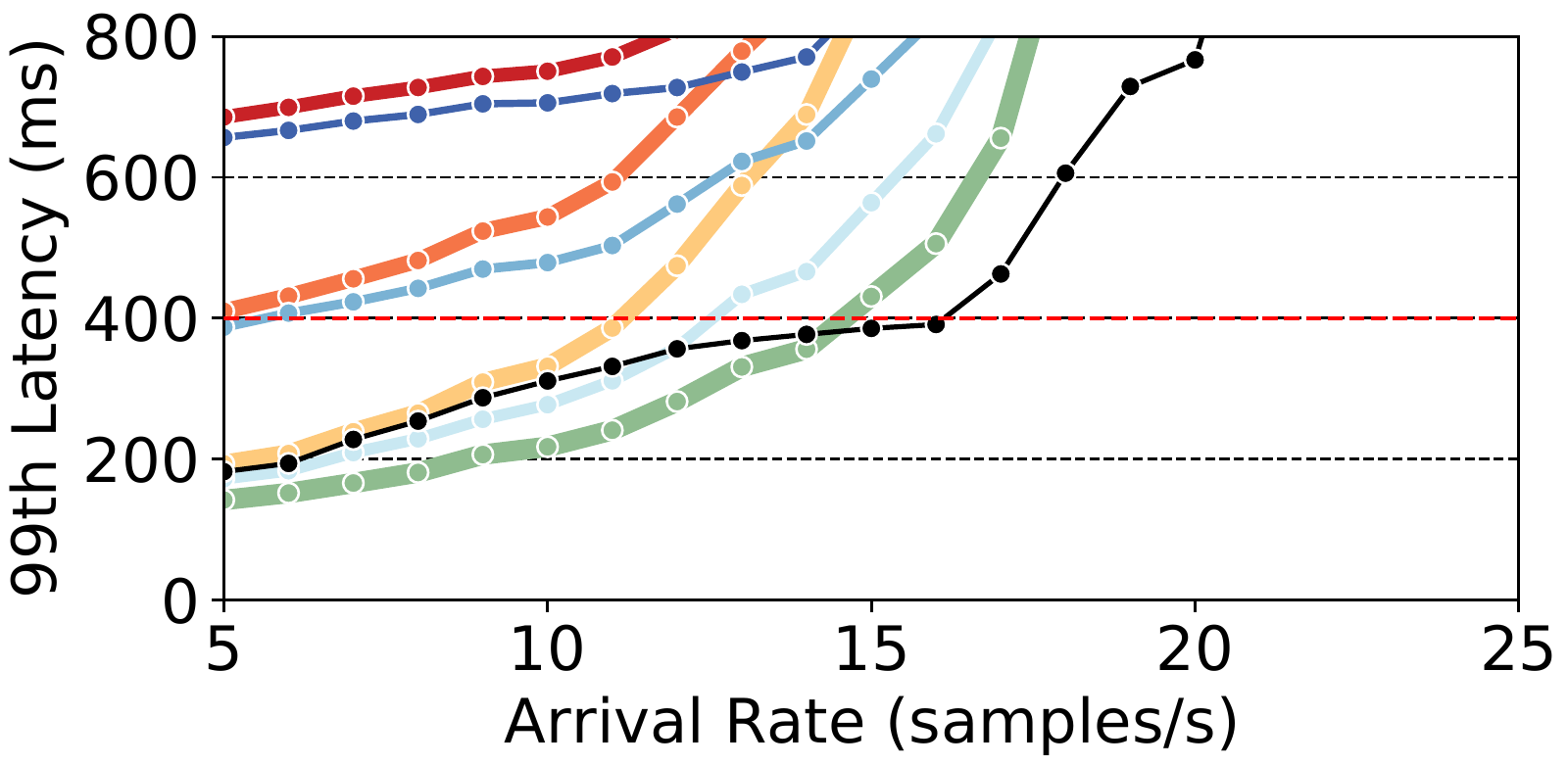}}
    \vspace{-1mm}
    \captionsetup{font=footnotesize,labelfont=bf}
    \caption{Comparison on 4-exit ResNet-50 on ZC706 under 400ms SLO.}
    \label{fig:perf_comp_zc706}
    \vspace{3mm}
\end{figure}

\section{Evaluation}
\label{sec:eval}

\noindent\textbf{Setup.}
We target Xilinx ZC706 hosting the mid-range Z7045 FPGA and Xilinx ZCU104 with the larger ZU7EV. All hardware designs were developed in Vivado HLS and clocked at 150 and 200 MHz for ZC706 and ZCU104, respectively. All designs were synthesised and placed-and-routed using Vivado Design Suite (v2019.2) and run on both boards. 
The Arm CPU was used to set up the AXI interfaces to the off-chip memory and run our scheduler. We measured performance via hardware counters and used \mbox{16-bit} fixed-point across all experiments. 

\vspace{1mm}
\noindent\textbf{Benchmarks.} We evaluate on two mainstream DNNs: ResNet-50 and Inception-v3, which constitute widely used  backbones across multiple downstream vision tasks. 
In the multi-exit setup, we adopt the design methodology~\cite{hapi2020iccad} of state-of-the-art hardware-aware early-exit models. The examined instances comprise three intermediate exits, placed equidistantly in terms of FLOPs on the corresponding frozen ImageNet-pretrained backbone. Similarly to relevant literature~\cite{hapi2020iccad, msdnet2017iclr}, softmax top-1 is employed as a metric for confidence, and the exit policy is tuned to minimise the workload while maintaining accuracy within 1.5 percentage points from the original backbone~\cite{branchynet2016icpr, flexdnn2020sec}. This optimisation led to a uniform confidence threshold of 0.8 across exits, yielding exit rates of 
{\small$\left<5.1\%, 16.9\%, 9.0\%, 69.0\%\right>$} and {\small$\left<14.5\%, 18.6\%, 22.2\%, 44.7\%\right>$}, and accuracies of 75.6\% and 75.8\%, for ResNet-50 and Inception-v3. 

\begin{table}[b]
    \vspace{2mm}
    \centering
    \captionsetup{font=small,labelfont=bf}
    \caption{\small Design Points and Resource Consumption.}
    \vspace{-1mm}
    \begin{tabular}{llll}
    \hline
    \multirow{2}{*}{\textbf{Model}} & \multirow{2}{*}{\textbf{Platform}} &\textbf{Design Point} & \textbf{Resource Utilisation} \\
    & & $\left<T_R,T_P,T_C\right>$ & [DSPs, BRAM, LUTs]\\
    \hline
    \multirow{2}{*}{ResNet-50}  &  ZC706 & {\small$\left< 4652,\phantom{1}7,128\right>$} & {\small [99.56, 99.96, 77]\%}  \\
      &  ZCU104 & {\small$\left< 6832,10,172\right>$} & {\small [99.53, 99.99, 78]\%}  \\
    \hline
     \multirow{2}{*}{Inception-v3} & ZC706 &  {\small$\left< 2742,\phantom{1}4,225\right>$} & {\small [100.0, 99.94, 75]\%} \\
      &  ZCU104 & {\small$\left< 6832,10,172\right>$} & {\small [100.0, 99.99, 79]\%} \\
    \hline
    \end{tabular}
    \label{tab:rsc_usage}
\end{table}

\vspace{1mm}
\noindent
\textbf{NPU DSE and Resource Usage.} Table~\ref{tab:rsc_usage} shows the designs generated by our DSE method (with $w_b$$=$$1$), together with their resource consumption. In addition to the processing engine, the Fluid Batching Engine's CU requires less than 0.5\% of LUTs. The FBCB consumes 0.57\% and 0.54\% of registers on ZC706 and ZCU104, while the larger FCBC of the deeper Inception-v3 uses 0.95\% and 0.90\%. 
The Stackable PEs instantiate demultiplexers and one LUT-based adder per two PEs, consuming less than 5.5\% and 6\% of the available LUTs for ResNet-50 and Inception-v3, respectively, on both devices. 


\vspace{1mm}
\noindent
\textbf{Baselines.} We compare against state-of-the-art (SOTA) batching approaches: 
\textit{i)}~single-sample execution (SERIAL), 
\textit{ii)}~FC-only batching (FC-AdaptB), 
\textit{iii)}~$R$ dim. batching (R-AdaptB), 
and \textit{iv)}~LazyBatching~\cite{lazybatching2021hpca}. 
Baselines \textit{ii)} and \textit{iii)} correspond to SOTA adaptive batching (AdaptB) systems~\cite{clipper2017NSDI,mark2019atc,nexus2019sosp,batch2020sc,equinox2021micro}.
\blue{Given the resource constraints of the target edge-grade platforms and DNN workloads, to support latency SLOs that make the system deployable, we need to limit batch size to 8. Indicatively, executing ResNet-50 on ZCU104 with $B$$=$$16$ yields an average latency of  277~ms that exceeds the 200-ms SLO examined below.} As such, we use $B_{\text{max}}$$=$$8$ across all baselines.
For FC and $R$, we set AdaptB's $T_{\text{timeout}}$ parameter to three different values: small (S), medium (M) and large (L) corresponding to a batch-forming waiting time equal to 5\%, 45\% and 95\% of the 99th percentile latency SLO. For the SLO-aware LazyBatching, we configure its scheduler with the respective SLO in each experiment.
For each baseline, we perform DSE using a batch size of 8 and select the highest performing NPU design for the target DNN-device pair.

\begin{figure}
    \centering
    \subfigure[Processing Rate vs Arrival Rate.]{\includegraphics[width=0.22\textwidth,trim={0cm 0cm 0cm 4.04cm},clip]{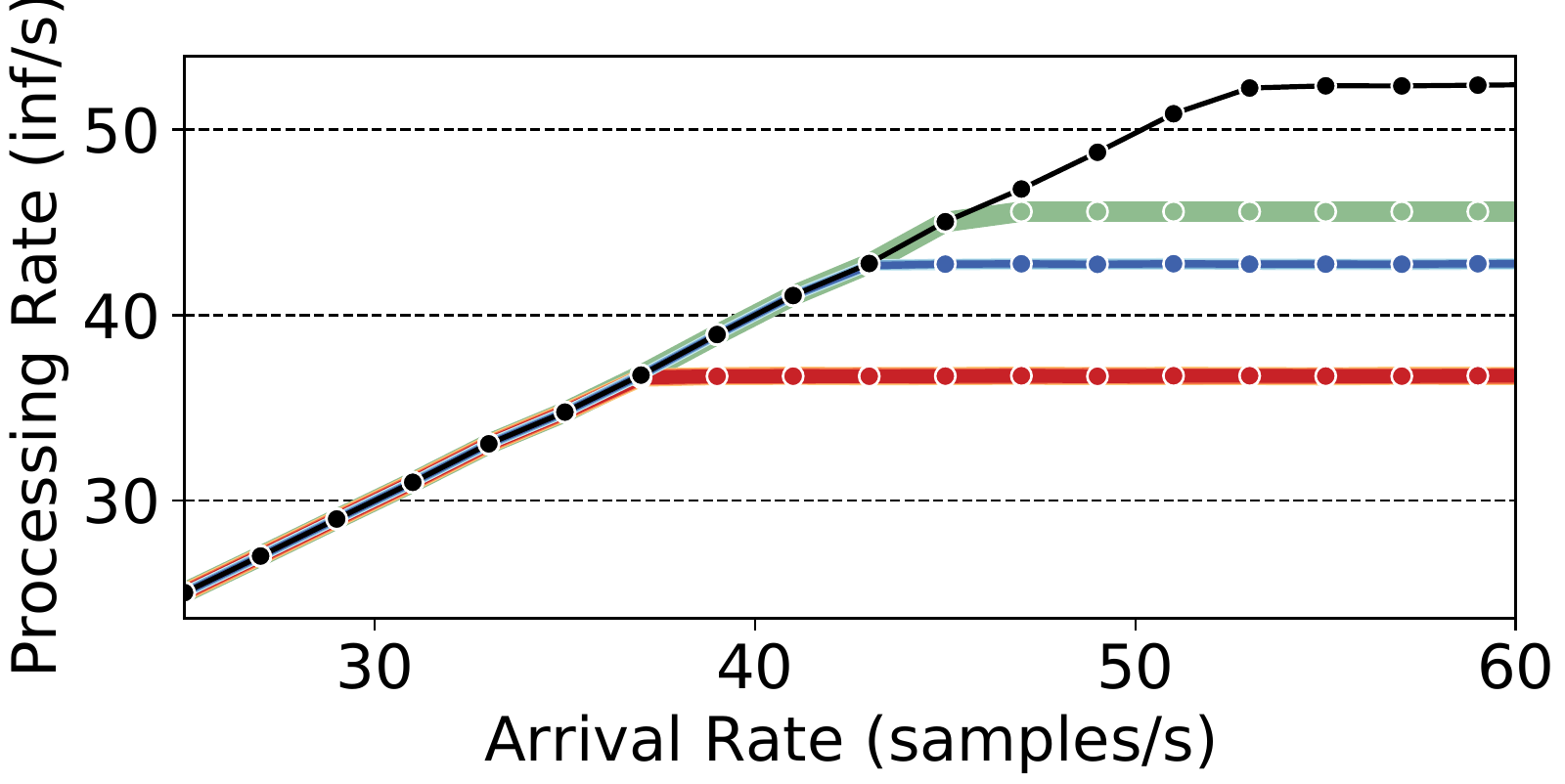}} 
    \subfigure[Avg. Latency vs Arrival Rate.]{\includegraphics[width=0.22\textwidth,trim={0cm 0cm 0cm 4.04cm},clip]{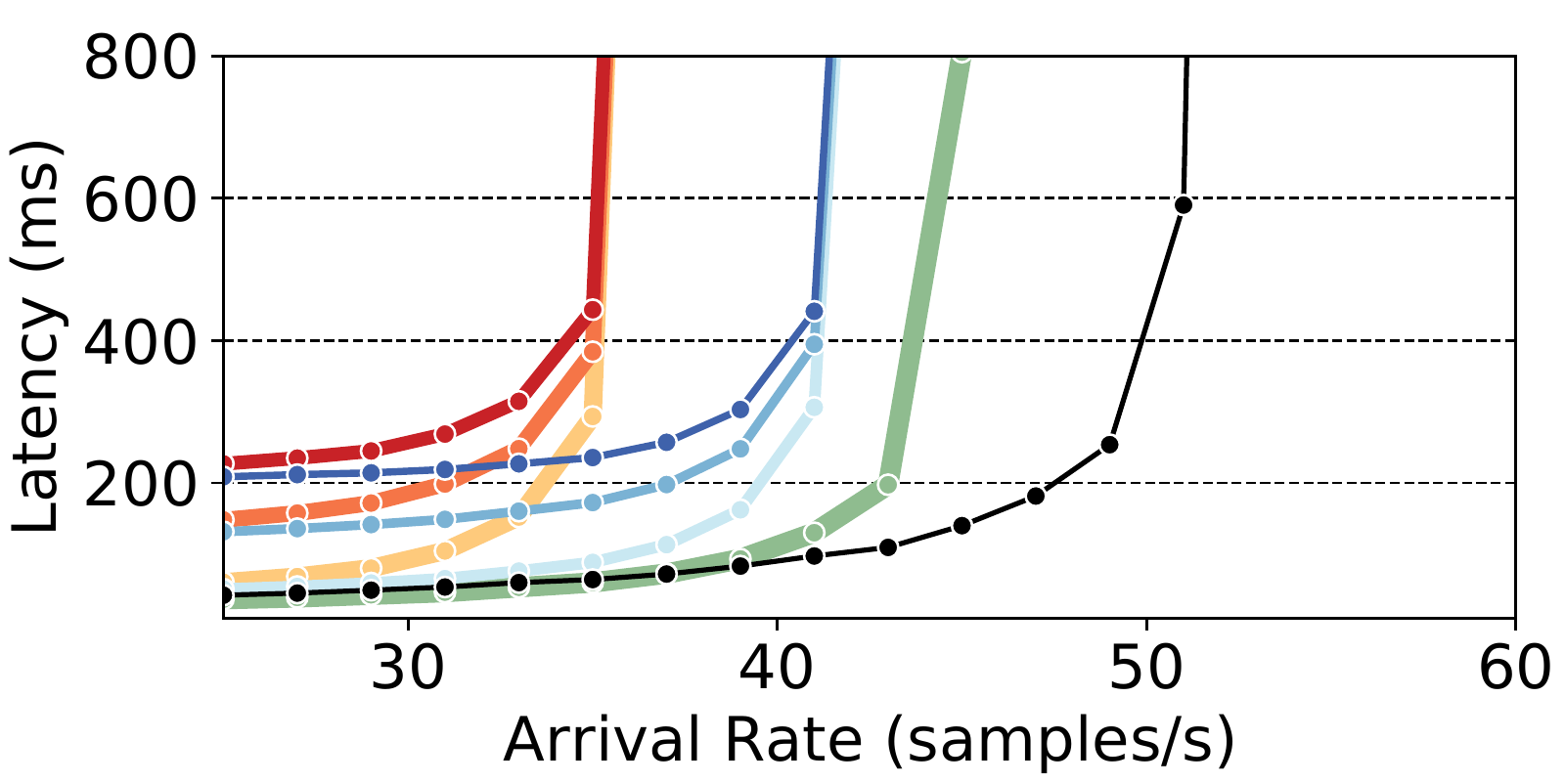}} 
    \subfigure[Utilisation vs Arrival Rate.]{\includegraphics[width=0.22\textwidth,trim={0cm 0cm 0cm 4.25cm},clip]{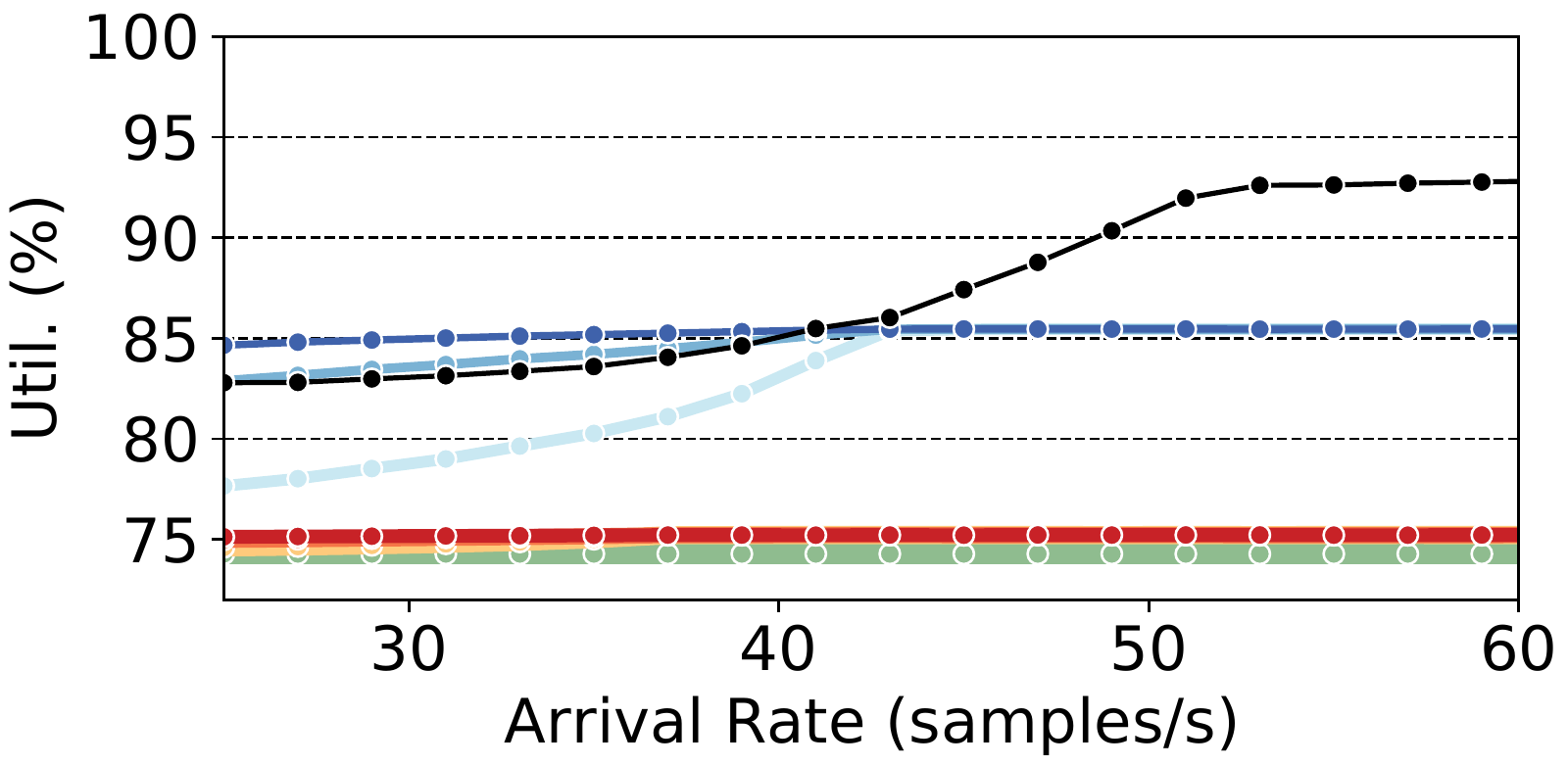}}
    \subfigure[Tail Latency vs Arrival Rate.]{\includegraphics[width=0.22\textwidth,trim={0cm 0cm 0cm 4.25cm},clip]{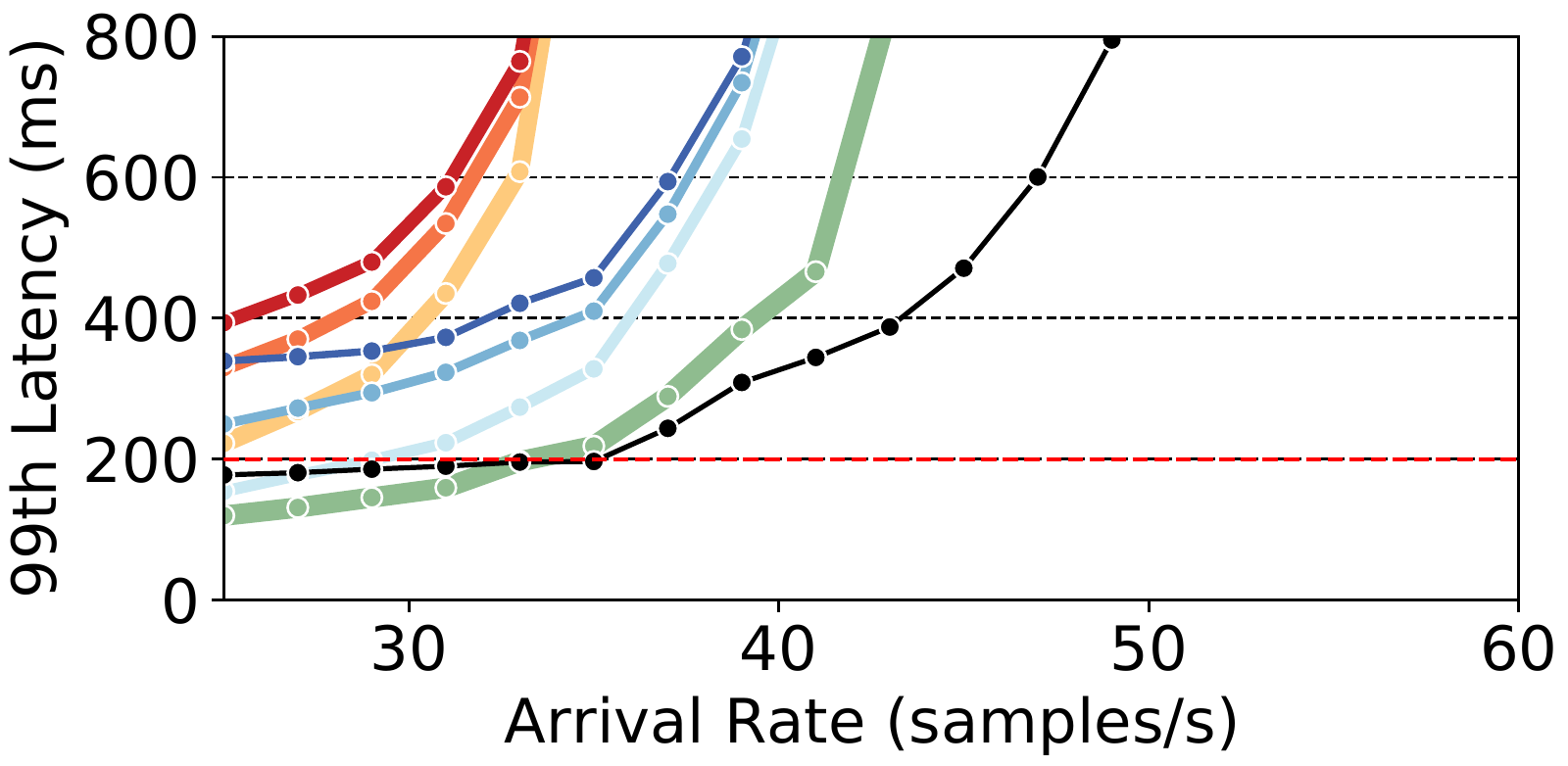}}
    \vspace{-1mm}
    \captionsetup{font=footnotesize,labelfont=bf}
    \caption{Comparison on 4-exit Inception-v3 on ZCU104 under 200ms SLO.}
    \label{fig:perf_comp_zcu104}
    \vspace{3mm}
\end{figure}

\vspace{1mm}
\noindent\textbf{Performance Comparison.}
To evaluate the performance and adaptability of our approach across traffic levels, we vary the request arrival rate between 5 and 25~samples/s and 20 and 60~samples/s for ResNet-50 (Fig.~\ref{fig:perf_comp_zc706}) and Inception-v3 (Fig.~\ref{fig:perf_comp_zcu104}), respectively. \blue{Following the MLPerf standard~\cite{mlperf_inf2020isca},} the arrival times of the input samples were generated following a Poisson distribution, with an expected rate equal to the specified samples/s. We set the 99th percentile latency SLO to 400~ms for ZC706 and to 200~ms for ZCU104. All experiments were run three times with different seeds and the average is reported.

\textbf{- Low-to-mid traffic.}
For slow-arriving samples, the large waiting windows of AdaptB-M and -L lead to excessive latency for batch forming, despite the higher utilisation due to large-batch processing. The small waiting windows of AdaptB-S, on the other hand, provide a better balance between utilisation and latency. SERIAL yields the lowest latency, but also the lowest utilisation. In contrast, our approach (FluidB) combines the merits of both approaches; it provides the user with SERIAL's QoE by meeting the SLO and achieving similar average latency, but also yields significantly higher utilisation (20.4\% average gain across the range 5 to 18 samples/s on ZC706 and 13.5\% average gain across the range 25 to 35 samples/s on ZCU104), by means of the flexible opportunistic batching of its scheduler. 


\textbf{- Mid-to-high traffic.}
For higher traffic, FC reaches a plateau in processing rate as it solely enables FC layers to benefit from batching. Similarly, $R$ reaches its limit, because batching uniformly only along the $R$ dimension cannot extract any additional performance. At the same time, all FC and $R$ variants gradually lead to excessive average and tail latency and constant utilisation. This can be attributed to the impact of early-exiting on batch size and the model-level batching of these approaches; as samples exit early, the effective batch size through the DNN is dynamically decreased. As FC and $R$ variants are not exit-aware and do not allow preemptions, they execute the rest of the DNN with smaller batch size and hence lower utilisation. This leaves input samples unnecessarily waiting in the queue, despite the extra batch room in the system.

Instead, FluidB exploits this extra room through its exit-aware scheduler that selectively preempts execution and merges new samples to form larger batches, thus processing the rest of the DNN with higher utilisation, while also benefiting from the enhanced mapping efficiency of all layers to the NPU. In addition, as the scheduler considers the SLO when making decisions, the average and tail latency are also kept under control. In particular, in traffic levels where there is enough slack from the SLO to perform a preemption, \textit{e.g.}~between 5 and 15~samples/s in Fig.~\ref{fig:perf_comp_zc706}d, the scheduler trades off a slightly higher tail latency \mbox{-still} without introducing violations- for a 20\% increase in NPU utilisation. Above 15~samples/s, FluidB provides significant gains in both average and tail latency, even over SERIAL. This can be attributed to the fact that under high traffic, incoming samples experience large waiting times. 
FluidB is able to better fill any space in the active batch through its flexible batching. As such, it cuts the waiting time and boosts the utilisation of the NPU, reaching close to 95\% for high traffic, far exceeding all alternatives. Finally, it sustains higher processing rate than other approaches, with above 20 and 55~inf/s on ZC706 and ZCU104, respectively.


\begin{figure}[t]
    \centering
    \vspace{-2mm}
    \subfigure[4-exit ResNet-50 - ZC706.]{\includegraphics[width=0.24\textwidth,trim={0cm 0cm 0cm 4.4cm},clip]{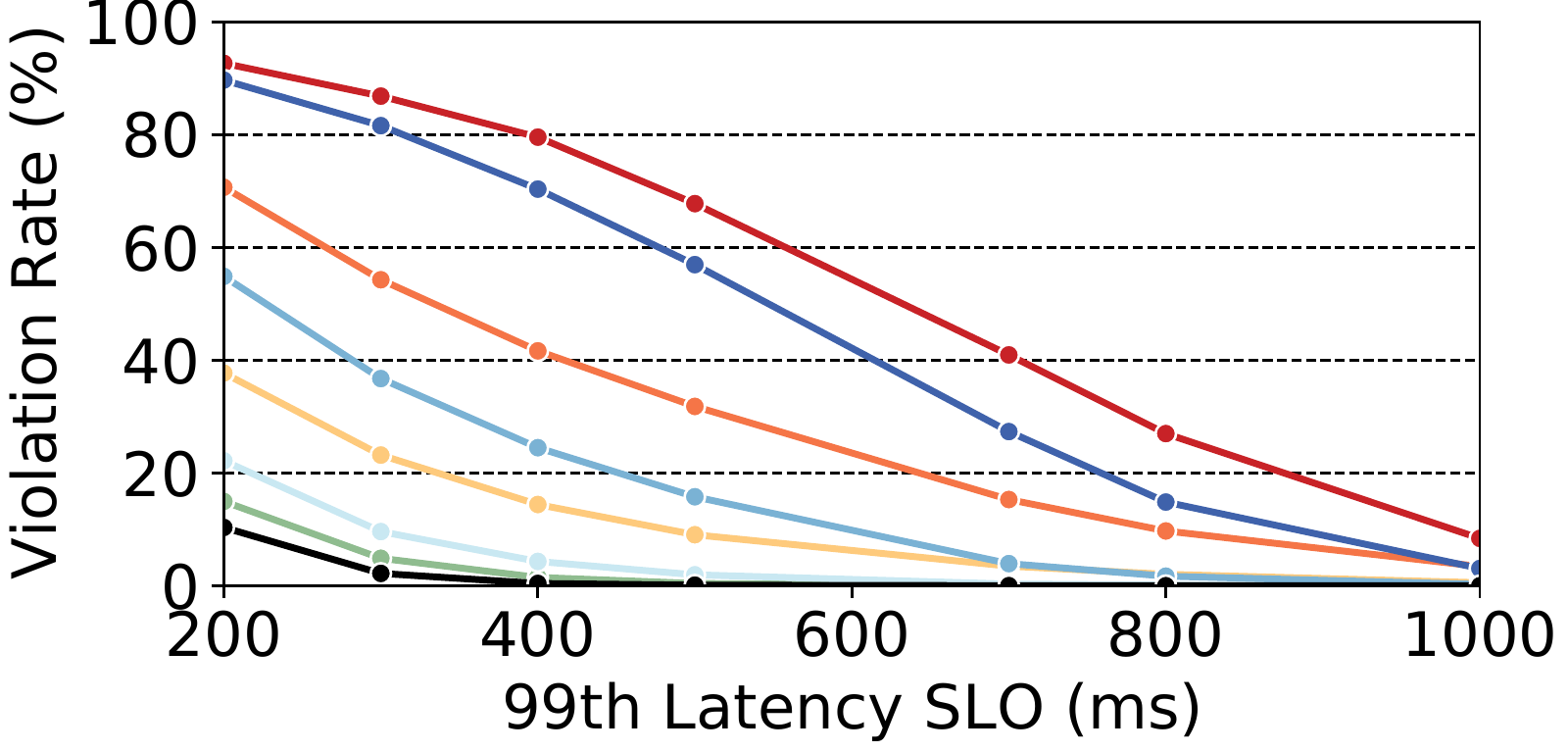}} 
    \subfigure[4-exit Inception-v3 - ZCU104.]{\includegraphics[width=0.24\textwidth,trim={0cm 0cm 0cm 4.4cm},clip]{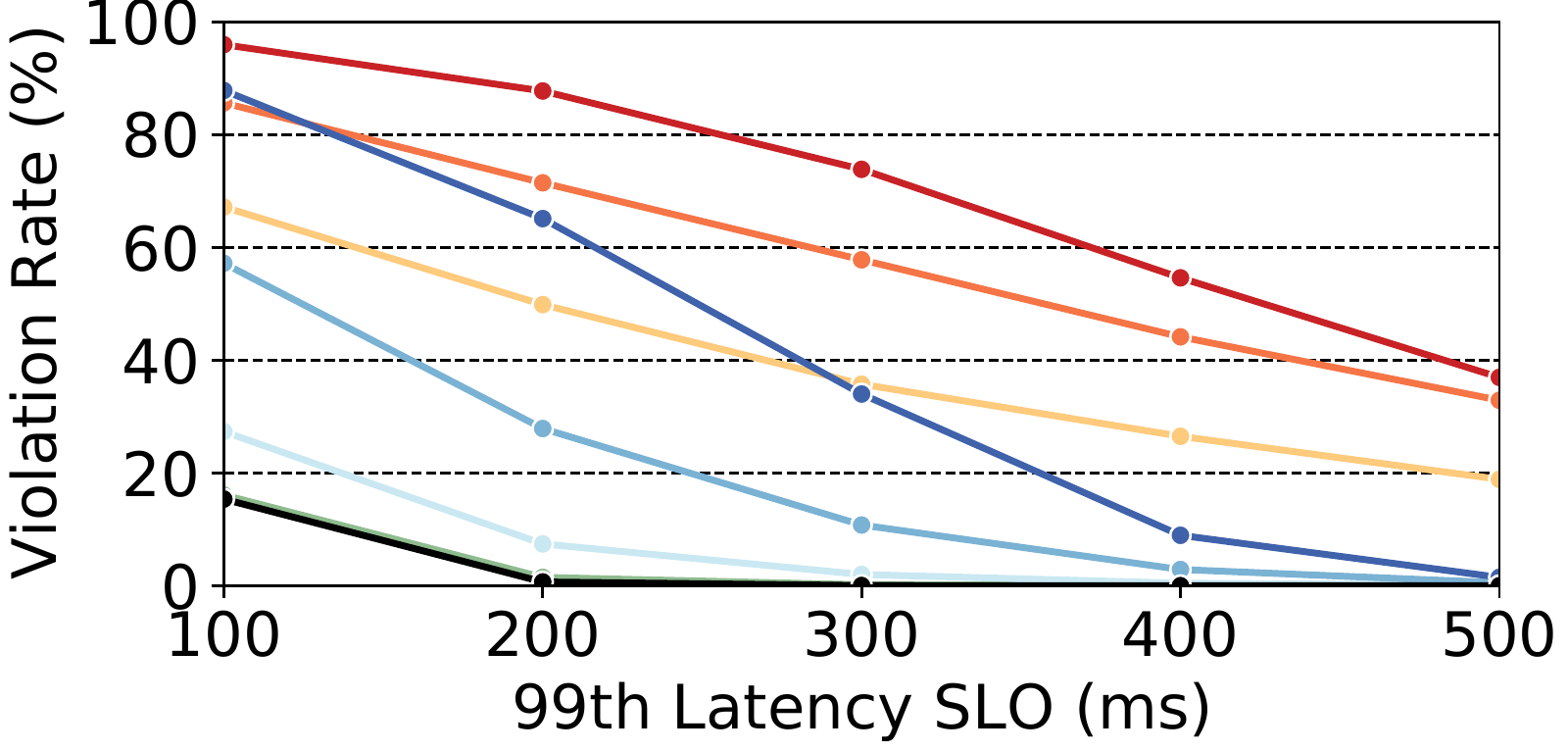}}
    \captionsetup{font=footnotesize,labelfont=bf}
    \vspace{-1mm}
    \caption{Violation rate as a function of tail latency SLO (x-axis) at 15 samples/s (left) and 35 samples/s (right). See Fig.~\ref{fig:perf_comp_zc706} for legend.}
    \label{fig:viol_rate}
\end{figure}

\vspace{1mm}
\noindent\textbf{Comparison to LazyBatching.}
Table~\ref{tab:lazy_batch} shows a comparison with LazyBatching~\cite{lazybatching2021hpca} as measured on ZC706 and ZCU104, with stringent tail latency SLOs and mid-to-high arrival rates. Targeting \mbox{ResNet-50}, our system achieves 1.43$\times$ and 2.5$\times$ lower average latency (1.89$\times$ geo. mean) while providing a significant SLO violation reduction of 13.2 percentage points (pp) and 27.1~pp (20.15~pp avg.), respectively.

LazyBatching introduces three sources of latency overhead. First, as it is exit-unaware, it invokes the preemption logic on every layer. As such, the associated latency is not fully amortised, affecting the average latency. Second, when the maximum batch size is reached, preemption is no longer considered, despite the fact that samples may exit early. Third, LazyBatching adopts a coarse and conservative method of estimating the latency of preemption, \textit{i.e.}~instead of considering the actual latency-throughput trade-off of batched execution, it approximates batched latency as the product of batch size and single-sample latency~\cite{lazybatching2021hpca}. This leads to an overestimation of the preemption overhead, with the system often deciding not to perform a preemption, even in cases where there is enough SLO slack and room in the batch due to early exiting. The overall effect is unnecessarily higher waiting time for many samples. Instead, our approach shows that accurately estimating the preemption overhead and incorporating exit-awareness into the scheduler are critical especially under high load and tight tail latency requirements, as they lead to well-amortised preemptions and significantly reduced SLO violations.

\begin{table}[t]
    \centering
    \captionsetup{font=small,labelfont=bf}
    \caption{\small Comparison with LazyBatching.}
    \vspace{-1mm}
    \resizebox{0.475\textwidth}{!}{
    \setlength{\tabcolsep}{1pt}
        \scriptsize
        \begin{tabu}{llll | rr | rr}
            \toprule
            & & & & \multicolumn{2}{c}{\textbf{LazyBatching}} & \multicolumn{2}{c}{\textbf{Fluid Batching}} \\
            \textbf{Model} & \textbf{Platform} & \textbf{Arrival Rate} & \textbf{SLO} & Avg. Lat. & Viol. Rate
            & Avg. Lat. & Viol. Rate \\ 
            \midrule
            ResNet-50 & ZC706 & 15 samples/s & 200 ms & 173 ms 
            & 23.53\% & 121 ms & 10.33\% \\
            
            & ZCU104 & 40 samples/s & 100 ms & 143 ms & 35.48\% 
            & \phantom{0}57 ms & 8.38\% \\
            
            
            \midrule
            
            Inception-v3 & ZC706 & 15 samples/s & 400 ms & 287 ms 
            & 15.94\% & 213 ms & 7.93\% \\
            
            & ZCU104 & 40 samples/s & 200 ms & 216 ms & 14.20\% 
            & \phantom{0}97 ms & 6.21\% \\
            
            \bottomrule
        \end{tabu}
    }
    \label{tab:lazy_batch}
    \vspace{3mm}
\end{table}

\vspace{1mm}
\noindent\textbf{Sensitivity to SLO.}
To assess robustness to different latency SLOs, we sweep the tail latency SLO and measure the violation rate of FluidB and the alternatives for the same arrival rate of 15 samples/s for ZC706 and 35 samples/s for ZCU104. As shown in Fig.~\ref{fig:viol_rate}, AdaptB variants experience significant violations even when the SLO is relaxed, \textit{i.e.}~to the right of the x-axis. FluidB achieve no violations unless the SLO is set to excessively low latencies considering the workload at hand,  
demonstrating its effectiveness even under stringent constraints. Compared to SERIAL, FluidB provides similar or fewer violations, at the added value of substantially improved NPU utilisation.

\begin{figure}[t]
    \centering
    \vspace{-2mm}
    \includegraphics[width=0.5\textwidth, trim={0cm 9.5cm 0cm 0cm},clip]{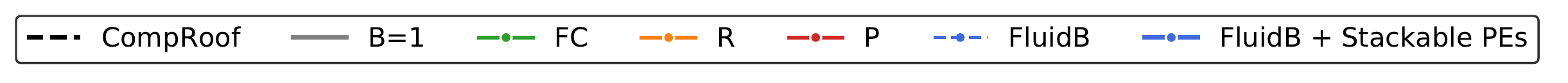}
    \subfigure[ResNet-50 - ZC706.]{\includegraphics[width=0.23\textwidth, trim={0cm 0cm 0cm 0cm},clip]{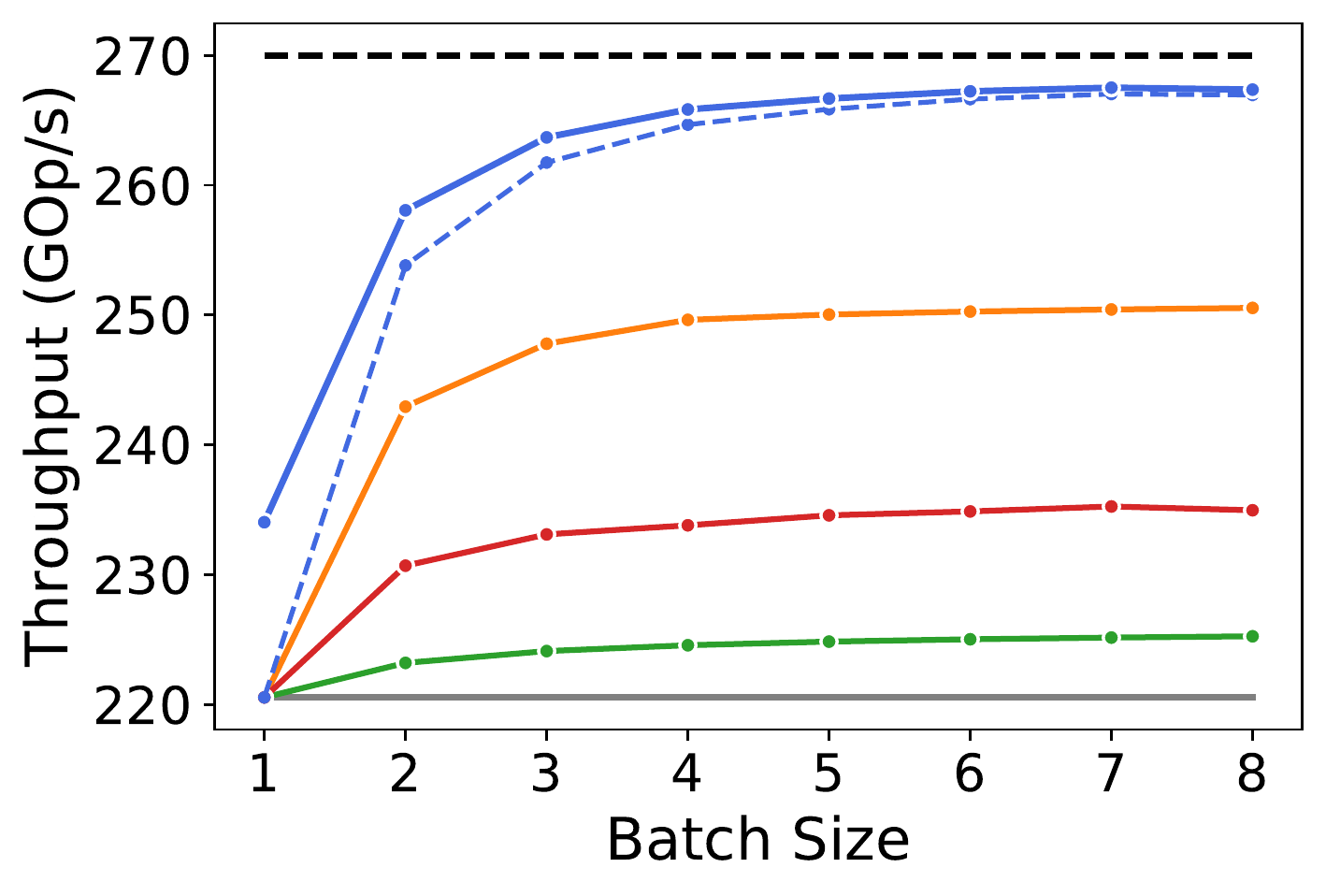}}
    \subfigure[Inception-v3 - ZCU104.]{\includegraphics[width=0.23\textwidth, trim={0cm 0cm 0cm 0cm},clip]{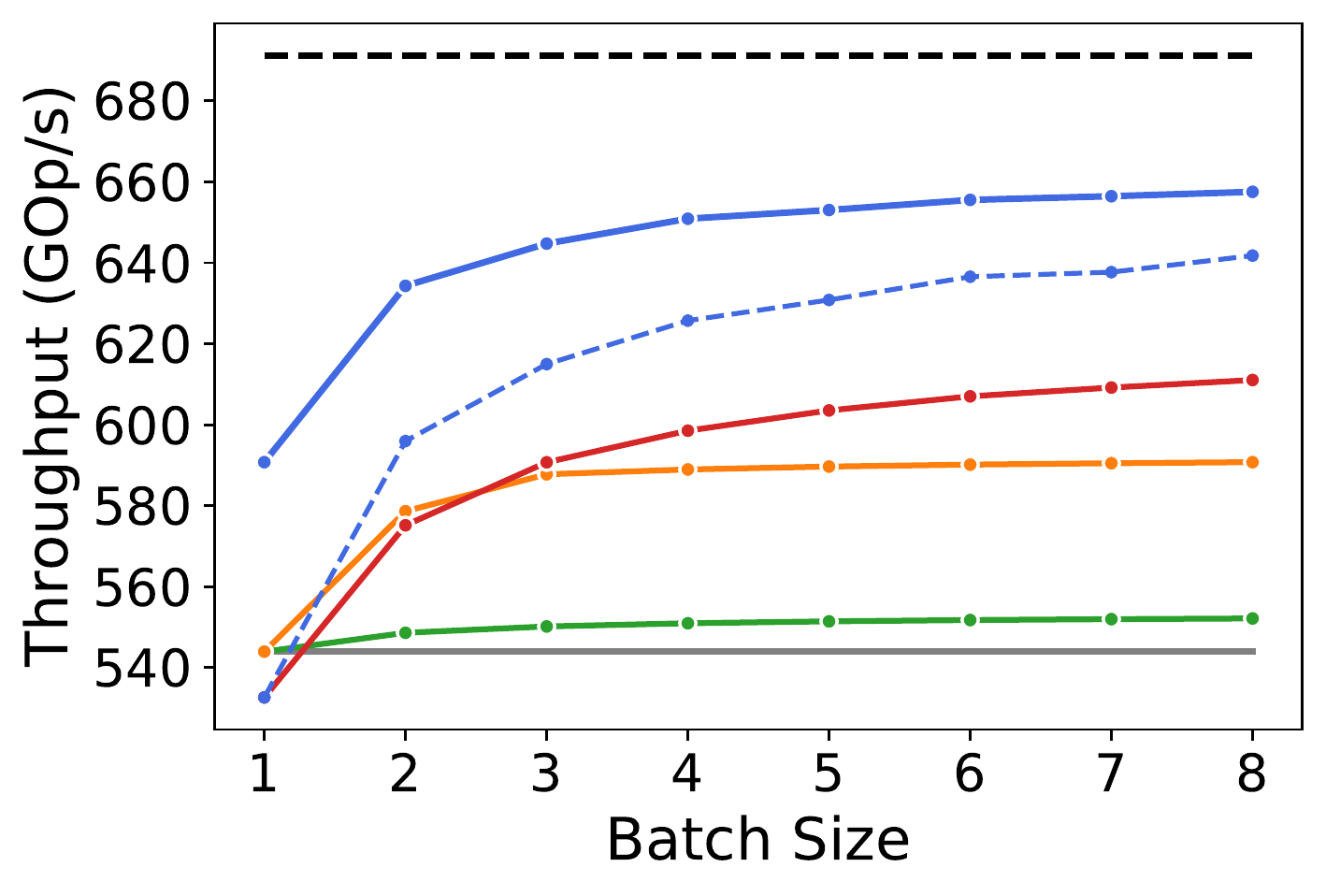}}
    \captionsetup{font=footnotesize,labelfont=bf}
    \vspace{-2mm}
    \caption{Comparison of Fluid Batching with status-quo batching techniques. The impact of Stackable PEs on performance is also ablated.}
    \label{fig:hw_perf}
    \vspace{3mm}
\end{figure}

\vspace{1mm}
\noindent\textbf{NPU Ablation.}
To evaluate the benefits of Fluid Batching and Stackable PEs on the NPU, we obtain several baselines by running DSE targeting the ResNet-50 and Inception-v3 backbones without early exits, considering multiple uniform batching strategies across layers and a static batch size during inference. Fig.~\ref{fig:hw_perf} shows the performance of the resulting designs across batch sizes.
We observe that Fluid Batching converges to significantly higher performance than all alternatives, even approaching the theoretical peak of the device in the case of ResNet-50 for $B$$>$$3$. Stackable PEs provide an additional performance boost that is more prominent for smaller batch sizes (incl. $B$$=$$1$). As such, the proposed methods act complementarily, offering pronounced gains across the spectrum.

\section{Discussion, Limitations \& Future Work}
Although our evaluation includes two diverse processing platforms and a broad range of request traffic and latency constraints, we have mainly evaluated the performance of the proposed system on convolutional neural networks. As early-exit mechanisms are being increasingly integrated into emerging Transformer architectures~\cite{fastbert2020acl,deebert2020acl,patient_bert2020neurips,edgebert2021micro}, an investigation on the applicability of the proposed techniques in this family of models constitutes a natural continuation of our work. Notably, the attention mechanism of Vision Transformers~\cite{vit_survey2023tnnls} comprises a mainly compute-bound and more regular workload~\cite{full_stack_opt_transformers2023arxiv} better fitting the GEMM structure of the examined accelerators, while demonstrating limited weight reuse between samples which leads to significantly less pronounced benefits from batching. However, exploiting different realisations of the architectural flexibility of the proposed NPU design (\textit{e.g.}~to dynamically trade intra- and inter-attention head parallelism at a layer granularity at run time to deal with varying input-sequence length), as well as the introduced exit-aware scheduler to make more informed preemption decisions and improve inference efficiency under such dynamic workloads, remains an open research question to be investigated in future work. 
Nonetheless, many Transformer architectures to date feature a mix of convolutional and attention layers~\cite{carion2020end,mobilevit2022iclr,li2022efficientformer} mapped on the same NPU. In such cases, Fluid Batching can equip the dimensionality of convolutional layers with further flexibility, leading potentially to a more efficient mapping to the hardware accelerator. 

Furthermore, in our prototype we have not allowed changes that significantly alter the accuracy of the original model. Towards introducing further flexibility, a future avenue would investigate the tuning of the confidence threshold -and thus the exit policy- at run time, as another means of optimising the execution under varying traffic.

Last, we have focused on edge-based settings, where the target platforms have limited computational capabilities and deployable configurations in terms of the maximum batch size supported. Future work would encompass scaling the proposed methods on cloud-grade setups, where batch sizes can be larger and traffic rates higher, but also offloading can be partial~\cite{distributed_branchynet2017icdcs,spinn2020mobicom,hastening2022tmc}. We also envision evaluating our approach on real-world traces of commercially available edge NPUs.

\section{Conclusion}
We have presented a framework for efficiently scheduling and serving multiple DNN inference requests of multi-exit models on edge NPUs. Despite the common belief that batch processing can be prohibitive for low-latency applications, we showed that dynamicity-aware preemptive scheduling yields the best of both worlds; the high utilisation of batching and the low latency of single-sample execution. Moreover, 
through hardware support for Fluid Batching, the attainable NPU utilisation 
is pushed beyond what was previously possible. Last, exit-awareness and accurate latency estimation leads to well-amortised preemptions, even under high load, and significantly fewer SLO violations.
%
We envision that this can pave the way towards research in hardware and dynamic DNN co-design.

\bibliographystyle{IEEEtran}
{\footnotesize
\bibliography{references}
}

\end{document}